\documentclass[12pt,letterpaper,notitlepage]{article}

\usepackage{tikz}
\usepackage{multicol}
\usepackage{tabularx}
\usepackage{booktabs}
\usepackage{times}
\usepackage{epsfig}
\usepackage{graphicx}
\usepackage{float}
\usepackage[margin={2cm,2cm}]{geometry}
\usepackage{ifthen}
\usepackage{subfig}
\usepackage{multirow}

\newboolean{arxiv}
\setboolean{arxiv}{true}

\usepackage[font={small}]{caption}

\usepackage{enumitem}
\setenumerate{itemsep=1pt,topsep=3pt}

\usepackage{boxedminipage}
\usepackage{wrapfig}

\newcommand{\cutsectionup}{}
\newcommand{\cutsectiondown}{}

\title{Exploring the structure of a real-time, arbitrary neural artistic stylization network}

\author{Golnaz Ghiasi\\
Google Brain\\
{\tt\small golnazg@google.com}
\and
Honglak Lee\\
Google Brain\\
{\tt\small honglak@google.com}
\and
Manjunath Kudlur\\
Google Brain\\
{\tt\small keveman@google.com}
\and
Vincent Dumoulin\\
MILA, Universit\'{e} de Montr\'{e}al\\
{\tt\small vi.dumoulin@gmail.com}
\and
Jonathon Shlens\\
Google Brain\\
{\tt\small shlens@google.com}
}

\begin{document}
\maketitle
\begin{abstract}


In this paper, we present a method which combines the flexibility of the neural algorithm of artistic style with the speed of fast style transfer networks to allow real-time stylization using any content/style image pair. We build upon recent work leveraging conditional instance normalization for multi-style transfer networks by learning to predict the conditional instance normalization parameters directly from a style image. The model is successfully trained on a corpus of roughly 80,000 paintings and is able to generalize to paintings previously unobserved. We demonstrate that the learned embedding space is smooth and contains a rich structure and organizes semantic information associated with paintings in an entirely unsupervised manner.

\end{abstract}

\section{Introduction}
\label{sec:intro}


{\it Elmyr de Hory} gained world-wide fame by forging thousands of pieces of
artwork and selling them to art dealers and museums \cite{irving1969}.
The forger's skill is a testament to the human talent and intelligence required to reproduce the artistic details of a diverse set of paintings.
In computer vision, much work has been invested in teaching computers to likewise
capture the artistic style of a painting with the goal of conferring
this style in arbitrary photographs in a convincing manner.

Early work in this effort in computer vision arose out of visual texture
synthesis. Such work focused on building non-parametric techniques for
``growing'' visual textures one pixel \cite{efros1999texture, wei2000fast}
or one patch \cite{efros2001image, liang2001real} at a time. Interestingly,
Efros et al. (2001) \cite{efros2001image} demonstrated that one may
transfer a texture to an
arbitrary photograph to confer it with the stylism of a drawing.
Likewise, Hertzmann et al. (2001) \cite{hertzmann2001image}
demonstrated a non-parametric
technique for imbuing an arbitrary filter to an image based on pairs of
unfiltered and filtered images.

In parallel to non-parametric approaches, a second line of research focused
on building parametric models of visual textures
constrained to match the
marginal spatial statistics of visual patterns \cite{julesz1962}. Early models
focused on matching the marginal statistics of multi-scale linear filter banks
\cite{portilla1999, freeman2011metamers}. In recent years, spatial image
statistics gleaned from intermediate features of state-of-the-art image
classifiers \cite{simonyan2014very}
proved superior for capturing visual textures \cite{gatys2015texture}.
Pairing a secondary constraint to preserve the content of an image -- as
measured by the higher level layers of the same image classification network
-- extended this idea to artistic style transfer \cite{gatys2015neural}
(see also \cite{gatys2}).

Optimizing an image or photograph to obey these constraints is computationally
expensive and contains no learned representation for artistic style.
Several research groups addressed this problem by building a secondary network, i.e.,
{\it style transfer network}, to explicitly learn the transformation from a
photograph a particular painting style
\cite{johnson2016perceptual, liwand2016, ulyanov2016texture}. Although this method confers
computational speed, much flexibility is lost: a single style transfer network
is learned for a single painting style and a separate style transfer network
must be built and trained for each new painting style.

\begin{figure*}
\begin{center}
\includegraphics[width=\linewidth]{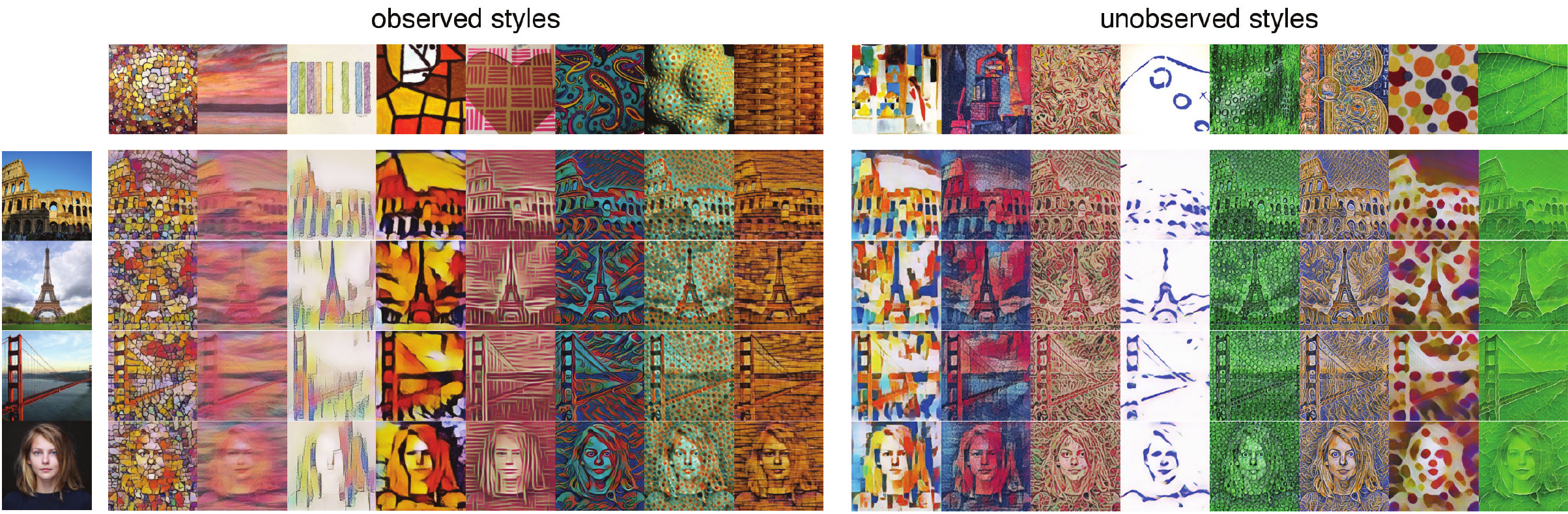}
\end{center}
\caption{Stylizations produced by our network trained on a large corpus of
paintings and textures. 
The left-most column shows four content images.
Left:
Stylizations from paintings in training set on paintings (4 left columns)
and textures (4 right columns). Right: Stylizations from paintings
never previously observed by our network.}
\label{fig:arbitrary}
\end{figure*}

Most crucially, by partitioning the style transfer problem customized for a specific style of painting,
these methods avoid the critical ability to learn a {\it shared}
representation across paintings. Recent work by Dumoulin et al.~\cite{dumoulin2016} demonstrated that the manipulation
of the normalization parameters was sufficient to train a single style transfer
network across 32 varied painting styles. Such a network
distilled the artistic style into a roughly 3000 dimensional space that is
regular enough to permit smooth interpolation between these painting styles.
Despite the promise, this model can cover only a limited number of styles and cannot generalize well to an unseen style. 
In this work, we extend these ideas further by building a style transfer network
trained on about 80,000 painting and 6,000 visual textures.
We demonstrate that this network can generalize to capture and transfer the artistic style of paintings
never previously observed by the system (see Figure~\ref{fig:arbitrary}). 
Our contributions in this paper include:
\begin{enumerate}
\item Introduce a new algorithm for fast, arbitrary artistic style transfer
trained on 80,000 paintings that can operate in real time on never previously observed paintings.
\item Represent all painting styles in a compact embedding space
that captures features of the semantics of paintings.
\item Demonstrate that training with a large number of paintings
uniquely affords the model the ability to predict styles never previously observed.
\item Embedding space permits novel exploration of artistic range of artist.
\end{enumerate}


\section{Methods}
\label{sec:methods}

\begin{figure}[t]
\begin{center}
\includegraphics[width=0.7\linewidth]{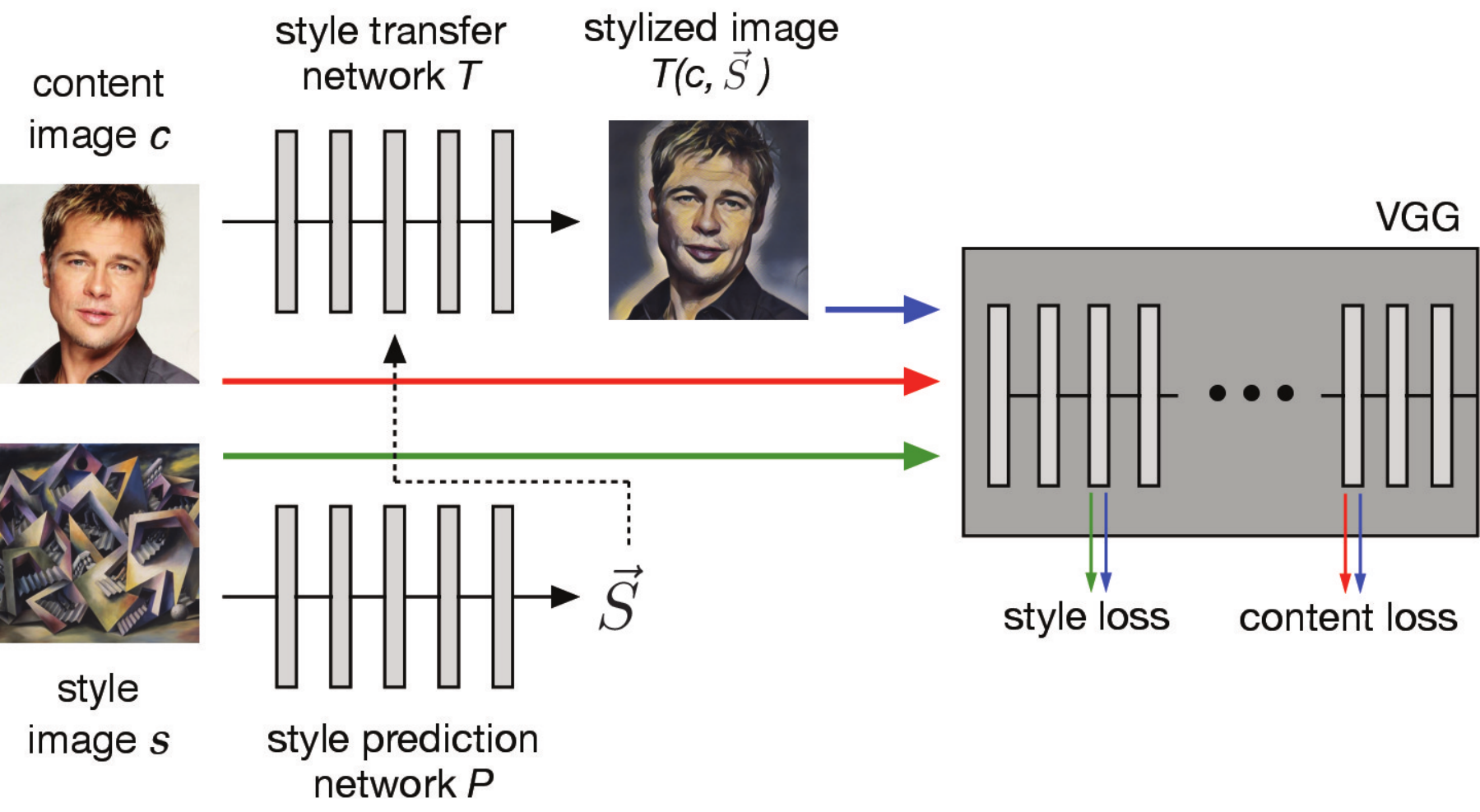}
\end{center}
\caption{Diagram of model architecture. The style prediction network $P(\cdot)$ predicts an embedding vector $\vec{S}$ from an input style image, which supplies a set of normalization constants for the style transfer network. The style transfer network transforms the
photograph into a stylized representation.
The content and style losses \cite{gatys2015neural} are derived from the distance
in representational space of the VGG image
classification network \cite{simonyan2014very}.
The style transfer network largely follows \cite{dumoulin2016} and the style
prediction network largely follows the Inception-v3 architecture \cite{szegedy2015}.}
\label{fig:model}
\end{figure}

Artistic style transfer may be defined as creating a stylized image $x$
from a content image $c$ and a style image $s$. Typically, the content image
$c$ is a photograph and the style image $s$ is a painting. A neural algorithm
of artistic style \cite{gatys2015neural} posits the content and style of an
image may be defined as follows:
\begin{itemize}
    \item Two images are similar in content if their high-level features as
        extracted by an image recognition system are close in Euclidean distance.
    \item Two images are similar in style if their low-level features as
        extracted by an image recognition system share the same
        spatial statistics.
\end{itemize}
The first definition is motivated by the observation that higher level features
of pretrained image classification systems are tuned to semantic information
in an image \cite{zeiler2014visualizing, johnson2016perceptual, inceptionism}.
The second definition
is motivated by the hypothesis that a painting style may be regarded as a
visual texture \cite{hertzmann2001image, efros2001image, gatys2015neural}.
A rich literature suggests that repeated motifs representative of a visual texture may be
characterized by lower-order spatial statistics \cite{julesz1962, portilla1999, freeman2011metamers}.
Images with identical lower-order spatial statistics appear perceptually
identical and capture a visual texture \cite{portilla1999, gatys2015texture, ulyanov2016instance, freeman2011metamers}.
Assuming that a visual texture is spatially homogeneous implies that
the lower-order spatial statistics may be represented by a Gram matrix expressing
the spatially-averaged correlations across filters within a given layer's representation
\cite{portilla1999, gatys2015texture, freeman2011metamers}.

The complete optimization objective for style transfer may be expressed as
\begin{equation}
\min_{x}\mathcal{L}_{c}(x,c)+\lambda_{s}\mathcal{L}_{s}(x,s)\label{eq:objective}
\end{equation}
where $\mathcal{L}_c(x, c)$ and $\mathcal{L}_s(x, s)$ are the content and style
losses, respectively and $\lambda_s$ is a Lagrange multiplier weighting the relative
strength of the style loss.
We associate lower-level and higher-level features as the activations
within a given set of lower layers $\mathcal{S}$ and higher layers $\mathcal{C}$
in an image classification network. The content and style losses are defined as
\begin{equation}
    \mathcal{L}_s(x,s) = \sum_{i \in \mathcal{S}}
        \frac{1}{n_i} \mid\mid \mathcal{G}[f_i(x)] - \mathcal{G}[f_i(s)] \mid\mid_F^2
\end{equation}
\begin{equation}
    \mathcal{L}_c(x,c) = \sum_{j \in \mathcal{C}}
        \frac{1}{n_j} \mid\mid f_j(x) - f_j(c) \mid\mid_2^2
\end{equation}
where $f_l(x)$ are the network activations at layer $l$, $n_l$ is the
total number of units at layer $l$ and $\mathcal{G}[f_l(x)]$ is the Gram matrix
associated with the layer $l$ activations. The Gram matrix is a square, symmetric
matrix measuring the spatially averaged correlation structure across the
filters within a layer's activations.

Early work focused on iteratively updating an image to synthesize
a visual texture \cite{portilla1999, freeman2011metamers, gatys2015texture}
or transfer an artistic style
to an image \cite{gatys2015neural}. This optimization procedure is slow and precludes any
opportunity to learn a representation of a painting style. Subsequent work
introduced a second network, a {\it style transfer network} $T(\cdot)$, to learn
a transformation from the content image $c$ to its artistically rendered version $\hat{x}$ (i.e., $\hat{x}=T(c)$) \cite{johnson2016perceptual, ulyanov2016instance, liwand2016}.
The style transfer network is a convolutional neural network formulated in the
structure of an encoder/decoder \cite{johnson2016perceptual, ulyanov2016instance}.
The training objective is the combination of style loss and content loss obtained by replacing $x$ in Eq.~\ref{eq:objective} with the network output $T(c)$.
The parameters of the style transfer network are trained by minimizing this objective using a corpus of photographic images as content. 
The resulting network may artistically render an image
dramatically faster, but a separate network must be learned for each
painting style.

Training a new network for each painting is wasteful because painting styles
share common visual textures, color palettes and semantics for parsing
the scene of an image. Building a style transfer network that shares
its representation across many paintings would provide a rich vocabulary for
representing any painting.
A simple trick
recognized in \cite{dumoulin2016} is to build a style transfer network as a
typical encoder/decoder architecture but specialize the normalization parameters
specific to each painting style. This procedure, termed {\it conditional
instance normalization}, proposes normalizing each unit's activation $z$ as
\begin{equation}
    \tilde{z} = \gamma_s \left (\frac{z - \mu}{\sigma} \right) + \beta_s
    \label{eq:normalization}
\end{equation}
where $\mu$ and $\sigma$ are the mean and standard deviation across the
spatial axes in an activation map \cite{ulyanov2016instance}.
$\gamma_s$ and $\beta_s$ constitute a linear transformation that specify the
learned mean ($\beta_s$) and learned standard deviation ($\gamma_s$)
of the unit. This linear transformation is unique to each painting style $s$.
In particular, the concatenation $\vec{S} = \{\gamma_s, \beta_s\}$ constitutes a
roughly 3000-d embedding vector representing the artistic style of a painting.
We denote this style transfer network as $T(\cdot, \vec{S})$. 
The set of all $\{\gamma_s, \beta_s\}$ across $N=32$ paintings constitute
0.2\% of the network parameters. 
Dumoulin et al.~\cite{dumoulin2016} showed that such a network provides
a fast stylization of artistic styles and the embedding
space is rich and smooth enough to allow users to combine the painting styles by \emph{interpolating} the learned embedding vectors of 32 styles.

Although an important step forward, this ``$N$-style network" is still limited
compared to the original optimization-based technique \cite{gatys2015neural}
because the network is limited to only work on the styles explicitly
trained on. The goal of this work is to extend this model to (1) train on
$N \gg 32$ styles and (2) perform stylizations for unseen painting styles never previously
observed. The latter goal is especially important because the degree to which
the network generalizes to unseen painting styles measures the degree to
which the network (and embedding space) represents the true breadth and diversity
of all painting styles.

In this work, we propose a simple extension in the form of
a
{\it style prediction network} $P(\cdot)$ that takes as input an \emph{arbitrary} style image $s$ and \emph{predicts} the embedding vector $\vec{S}$ of normalization constants, as illustrated in Figure~\ref{fig:model}. 
The crucial advantage of this approach is that the model can generalize to an unseen style image by predicting its proper style embedding at test time. 
We employ a pretrained Inception-v3 architecture
\cite{szegedy2015} and compute the mean across each activation channel of
the Mixed-6e layer which returns a feature vector with the dimension of 768. Then we apply two
fully connected layers on top of it to predict the final embedding $\vec{S}$.
The first fully connected layer is purposefully constructed to contain 100 units
which is substantially smaller than the dimensionality of $\vec{S}$ in order to compress the representation.
We find it sufficient to jointly train the style prediction network $P(\cdot)$ and style transfer network $T(\cdot)$ on a large corpus of photographs and paintings.

A parallel work has proposed another method for fast, arbitrary style transfer in
real-time using deep networks \cite{huang2017}. Briefly, Huang et al (2017)
employ the same transformation (Equation \ref{eq:normalization}) to normalize activation
channels, however they calculate $\gamma_s$ and $\beta_s$ as the mean and
standard deviation across the spatial axes of an encoder network applied
to a style image. Although the style transformation is simpler,
it provides a fixed heuristic mapping from style image to normalization parameters, 
whereas our method learns the mapping from the style image to style parameters directly.
Our experimental results indicate that the increased
flexibility achieves better objective values in the optimization.

\section{Results}
\label{sec:results}

We train the style prediction network $N(\cdot)$ and style transfer network
$T(\cdot)$ on the ImageNet dataset as a corpus of training content images
and the Kaggle {\it Painter By Numbers} (PBN) dataset\footnote{
https://www.kaggle.com/c/painter-by-numbers}, consisting of 79,433 labeled
paintings across many genres, as a corpus of training style images.
Additionally, we train the model when
{\it Describable Textures Dataset} (DTD) is used as the corpus of training style images.
This dataset consists of 5,640 images labeled across 47 categories \cite{cimpoi14describing}.
In both cases, we agument the training style images. We randomly flip, rescale, crop the images
and change the hue and contrast of them.
We present our results on both training style dataset.

\subsection{Trained network predicts arbitrary painting and texture styles.}
\label{sec:arbitrary}

Figure \ref{fig:arbitrary} (left) shows stylizations from the network
trained on the DTD and the PBN datasets. The figure highlights a number of stylizations across a few photographs. 
We note that the networks
were trained jointly and unlike previous work
\cite{dumoulin2016, johnson2016perceptual}, it was unnecessary to
select a unique Lagrange multiplier $\lambda_s$ for each painting style.
That is, a single weighting of style loss suffices to produce reasonable
results across all painting styles and textures.

Importantly, we employed the trained networks to predict a stylization
for paintings and textures never previously observed by the network
(Figure \ref{fig:arbitrary}, right). Qualitatively, the artistic
stylizations appear to be indistinguishable from stylizations produced
by the network on actual paintings and textures the network was trained
against.
We took this as an encouraging sign that the network learned a general
method for artistic stylization that may be applied for arbitrary
paintings and textures.
In the following sections we quantify this behavior
and measure the limits of this generalization.

\subsection{Generalization to unobserved paintings.}
\label{sec:generalization}

\begin{figure*}[t]
\begin{center}
\includegraphics[width=\linewidth]{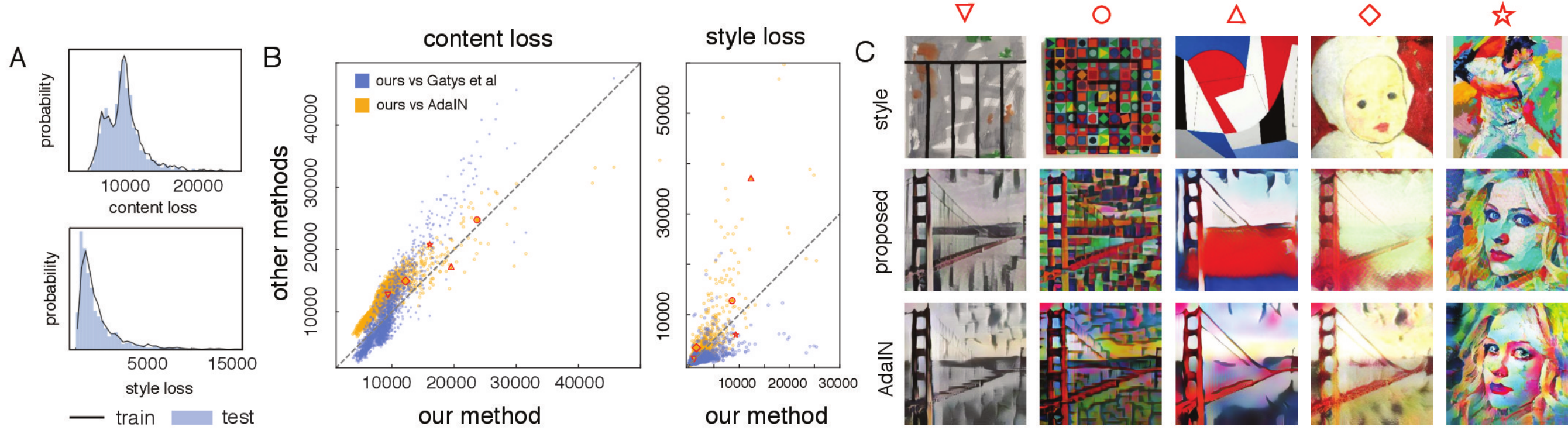}
\end{center}
\caption{Generalization to unobserved painting styles. A. Distribution of style and content
loss for stylization using observed and unobserved paintings from PBN training set.
B. Comparison of style and content
loss between proposed method, direct optimization \cite{gatys2015neural} (blue)
and AdaIN \cite{huang2017} (yellow). C. Sample images demonstrating stylization
applied between proposed method and AdaIN \cite{huang2017} for selected points in panel B.}
\label{fig:generalization}
\end{figure*}


Figure \ref{fig:arbitrary} indicates that the model is able to predict stylizations
for paintings and textures never previously observed that are qualitatively
indistinguishable from the stylizations on trained paintings and textures.
In order to quantify this observation, we train a model on the PBN dataset
and calculate the distribution of
style and content losses across 2 photographs for 1024 observed painting
styles (Figure \ref{fig:generalization}A, black)
and 1024 unobserved painting styles (Figure \ref{fig:generalization}A, blue).
The distribution of losses for observed styles (style: mean = $2.08e4 \pm 2.50e4$;
content: mean = $8.92e4 \pm 3.13e4$)
is largely similar to the distribution across unobserved styles
(style: mean = $1.95e4 \pm 3.73e4$; content: mean = $8.94e4\pm 3.55e4$).
This indicates that the method performs stylizations
on observed paintings with nearly equal fidelity as measured by the model
objectives for unobserved styles.
Importantly, if we train the model on a distinct but rich visual textures dataset
(DTD) and test the stylizations on unobserved paintings from PBN, we find
that the model produces similar artistic stylizations both quantitatively 
(style: mean = $2.67e4\pm 6.49e4$; content: mean = $8.76e4\pm 3.55e4$) and 
qualitatively (in terms of visual inspection). 
Due to space constraints, we provide detailed analysis in the supplementary material.


We next asked how well
the learned networks perform on unobserved painting styles when compared to the original
optimization-based
method \cite{gatys2015neural}. Figure \ref{fig:generalization}B plots the content and style loss
objectives for our proposed method (x-axis) and \cite{gatys2015neural} (blue points).
Note that even though \cite{gatys2015neural} directly optimizes for these two objectives,
the proposed method obtains content and style losses that are comparable
(style: $1.95e4$ vs $1.12e4$;
content: $8.94e4$ vs $9.09e4$). These
results indicate that the learned representation is able to achieve an objective
comparable to one obtained by direct optimization on the image itself.
%

We additionally compared our proposed method against a parallel work to
perform fast, arbitrary stylization termed {\it AdaIN} \cite{huang2017}.
We found that our
proposed method achieved lower content and style loss.
Specifically, 
(style: $1.95e4$ vs $2.56e4$;
content: $8.94e4$ vs $12.3e4$). 
In addition, paired t-test showed that these differences are statistically significant (style: p-value of $1.9\times 10^{-9}$ with t-statistic of $-6.04$; content: p-value of $0.0$ with t-statistic of $-91.9$), 
indicating that our proposed model achieved consistently better dual objectives
(Figure \ref{fig:generalization}B, yellow points). 
See Figure \ref{fig:generalization}C for a comparison of each method.

\begin{wrapfigure}{r}{0.35\textwidth}
  \vspace*{-0.3in}
  \begin{center}
    \includegraphics[width=0.3\textwidth]{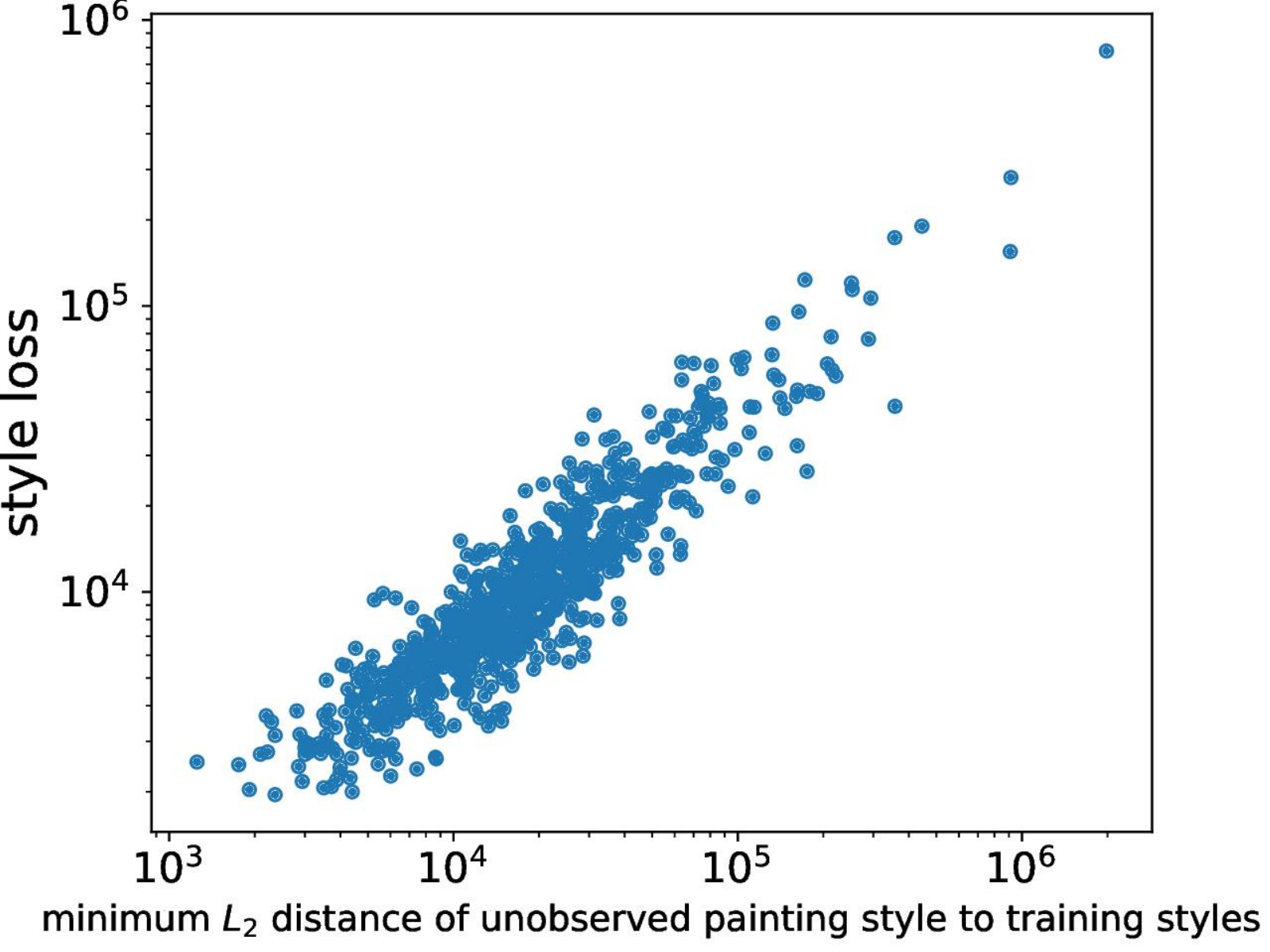}\\
  \end{center}
  \caption{Ability to generalize vs. proximity to training examples}
\label{fig:generalization-explain}
\end{wrapfigure}

Figure~\ref{fig:generalization-explain} shows how the generalization ability of the model (measured in terms of style loss) is related to the proximity to training examples. Specifically, we plot style loss on unobserved paintings versus the minimum $L_2$ distance between the Gram matrix of unobserved painting and the set of all Gram matrices in the training dataset of paintings. The plot shows clear positive correlation ($r^2 = 0.9$), which suggests that our model achieves lower style loss when the unobserved image is similar to some of the training examples in terms of the Gram matrix. More discussion of this figure is found in the supplementary material.

\subsection{Scaling to large numbers of paintings is critical for generalization.}
\label{sec:scaling}

\begin{figure*}[t]
\begin{center}
\includegraphics[width=0.9\linewidth]{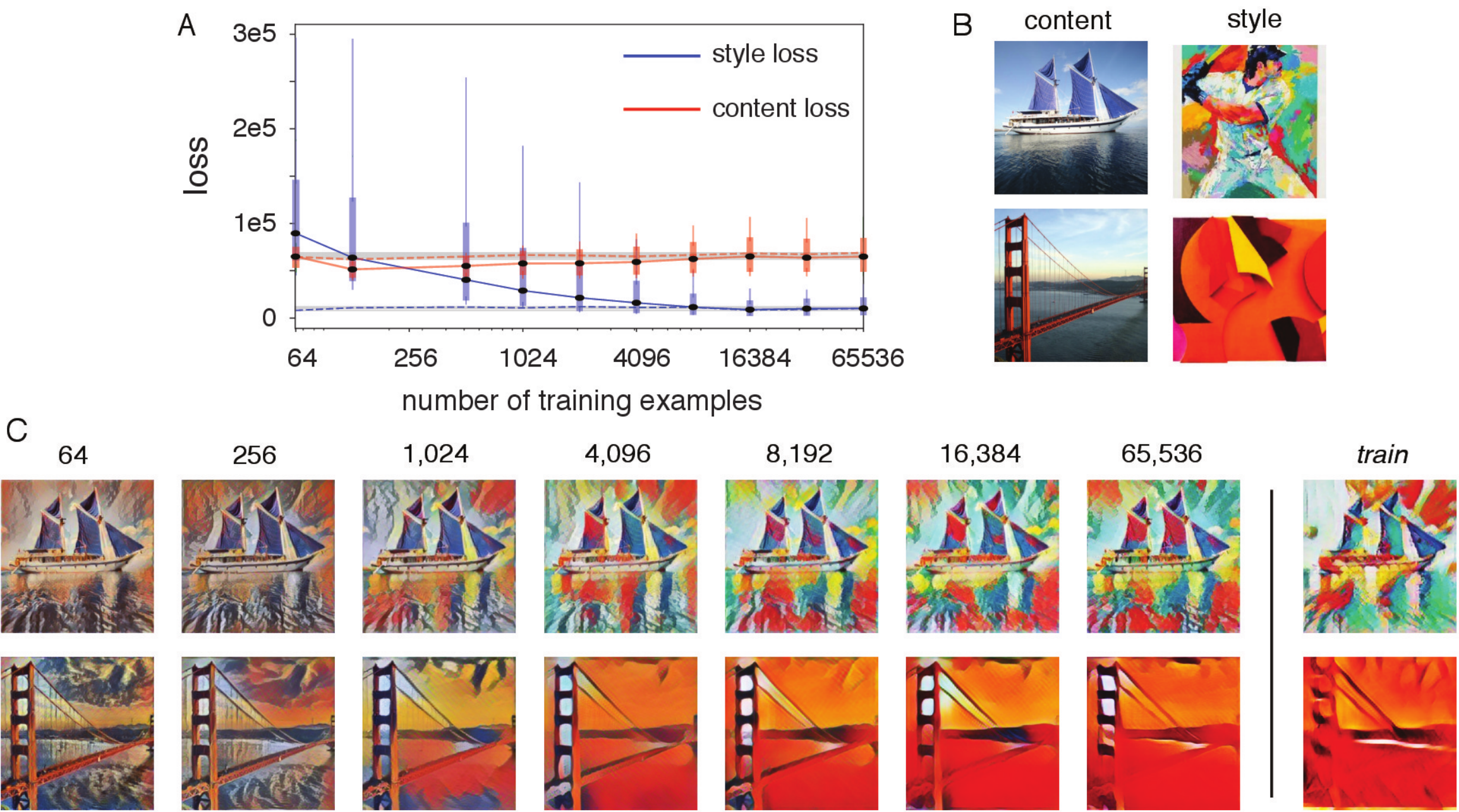}
\end{center}
\caption{Training on a large corpus of paintings is critical for generalization.
A. Distribution of style and content loss for stylizations applied to unseen painting styles
for proposed method trained on increasing numbers of painting styles.
Solid line indicates median with box showing $\pm 25\%$ quartiles and
whiskers indicating $10\%$ and $90\%$ of the cumulative distributions.
Dashed line and gray region indicate the mean and range of the
corresponding losses for training images.
Three sample pairs of content and style images (B) and the resulting stylization
with the proposed method as the method is trained on increasing numbers of paintings (top number).
For comparison, final column in (B) highlights stylizations for a model trained
explicitly on the these styles.}
\label{fig:scaling}
\end{figure*}

A critical question we next asked was what endows these networks with the ability
to generalize to paintings not previously observed. We had not observed this ability
to generalize in previous work \cite{dumoulin2016}. A simple hypothesis is that
the generalization is largely due to the fact that the model is trained with a
far larger number of paintings than previously attempted. To test this hypothesis,
we trained style transfer and style prediction networks with
increasing numbers of example painting styles without data augmentation.
Figure \ref{fig:scaling}A reports the distribution of content and style loss
on unobserved paintings for increasing numbers of paintings.

First, we asked whether the model is better able to stylize photographs
based on paintings in the training set by dint of having trained on
larger numbers of paintings. Comparing left-most and right-most points of
the dashed curves in Figure \ref{fig:scaling}A for the content and style loss
indicate no significant difference. Hence, the quality of the stylizations
for paintings in the training set do not improve with increasing numbers of paintings.

We next examined how well the model is able to generalize when trained on
increasing numbers of painting styles.
Although the content
loss is largely preserved in all networks, the distribution of style losses is
notably higher for unobserved painting styles and this distribution does not asymptote
until roughly 16,000 paintings. Importantly, after roughly 16,000 paintings
the distribution of content and style loss roughly match the content and style loss
for the trained painting styles. Figure \ref{fig:scaling}B shows three pairings
of content and style images that are unobserved in the training data set and
the resulting stylization as the model is trained on increasing number of paintings
(Figure \ref{fig:scaling}C). Training on a small number of paintings produces poor
generalization whereas training on a large number of paintings produces
reasonable stylizations on par with a model explicitly trained on this painting style.


\subsection{Embedding space captures semantic structure of styles.}
\label{sec:embedding}

The style transfer model represents all paintings and textures in a style embedding vector $\vec{S}$
that is 2758 dimensional. The style prediction network predicts $\vec{S}$ from a lower dimensional
representation (i.e., bottleneck) containing only 100 dimensions.

Given the compressed representation
for all artistic and texture styles, one might suspect that the network would automatically organize
the space of artistic styles in a perceptually salient manner. Furthermore, the degree to which this
unsupervised representation of artistic style matches our semantic categorization of paintings.

We explore this question by qualitatively examining the low dimensional representation for style
internal to the style prediction network.
A 100 dimensional space is too large to visualize, thus we
employ the t-SNE dimensional reduction technique to reduce the representation to two
dimensions \cite{maaten2008visualizing}.
Note that t-SNE will necessarily distort the representation significantly in order compress
the representation to small dimensionality, thus we restrict our analysis to qualitative description.

\begin{figure*}
\begin{center}
\includegraphics[width=\linewidth]{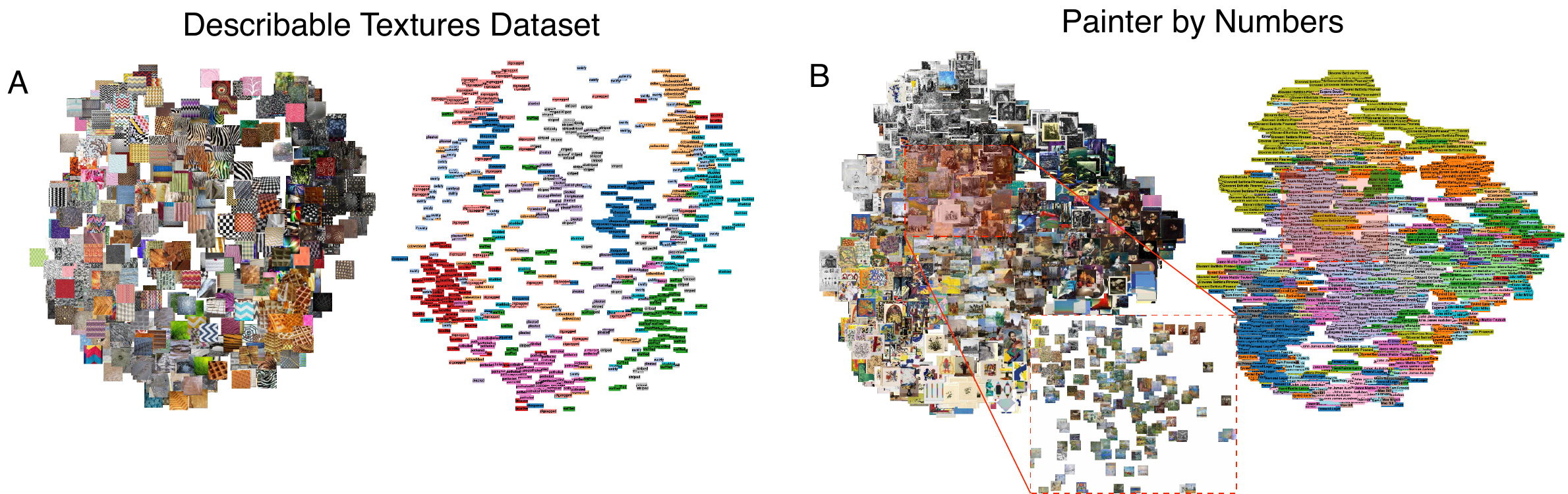}
\end{center}
\caption{Structure of a low-dimensional representation of the embedding space.
A: Two-dimensional representation using t-SNE for 800 textures
\cite{cimpoi14describing}
across
10 human-labeled categories. Right is the same as previous but texture
replaced with a human annotated label.
B: Same as previous but with Painting by Numbers dataset across
for 3768 paintings across 20 labeled artists.
Note the zoom-in highlighting a localized region
of embedding space representing Monet paintings. 
Please zoom-in for details.
}
\label{fig:embedding}
\end{figure*}

Figure \ref{fig:embedding}A (left) shows a two-dimensional t-SNE representation on a subset of 800 textures 
across 10 human-labeled categories. One may identify that regions
of the embedding space cluster around perceptually similar visual textures: the bottom-right contains
a preponderance of waffles; the middle contains many checkerboard patterns; top-center contains
many zebra-like patterns.
Figure \ref{fig:embedding}B (left) shows the same representation for a subset of 3768 paintings across
20 artists. Similar clustering behavior may be observed across colors and spatial structure as well.

The structure of the low dimensional representation does not just contain visual similarity but also
reflect semantic similarity.
To highlight this aspect, we reproduce the t-SNE plot but replace the individual images
with a human label (color coded). For the visual texture embedding (Figure \ref{fig:embedding}A) we display
a metadata label associated with each human-described texture. For the painting embedding
(Figure \ref{fig:embedding}B) we display the name of the artist for each painting.
Interestingly, we find that resides a region of the low-dimensional space that contains a large
fraction of Impressionist paintings by Claude Monet (Figure \ref{fig:embedding}B, magnified in inset).
These results suggest that the style prediction network has learned a representation for artistic styles
that is largely organized based on our perception of visual and semantic similarity 
without any explicit supervision.


\subsection{The structure of the embedding space permits novel exploration.}
\label{sec:exploration}

\begin{figure*}[t]
\begin{center}
\includegraphics[width=0.9\linewidth]{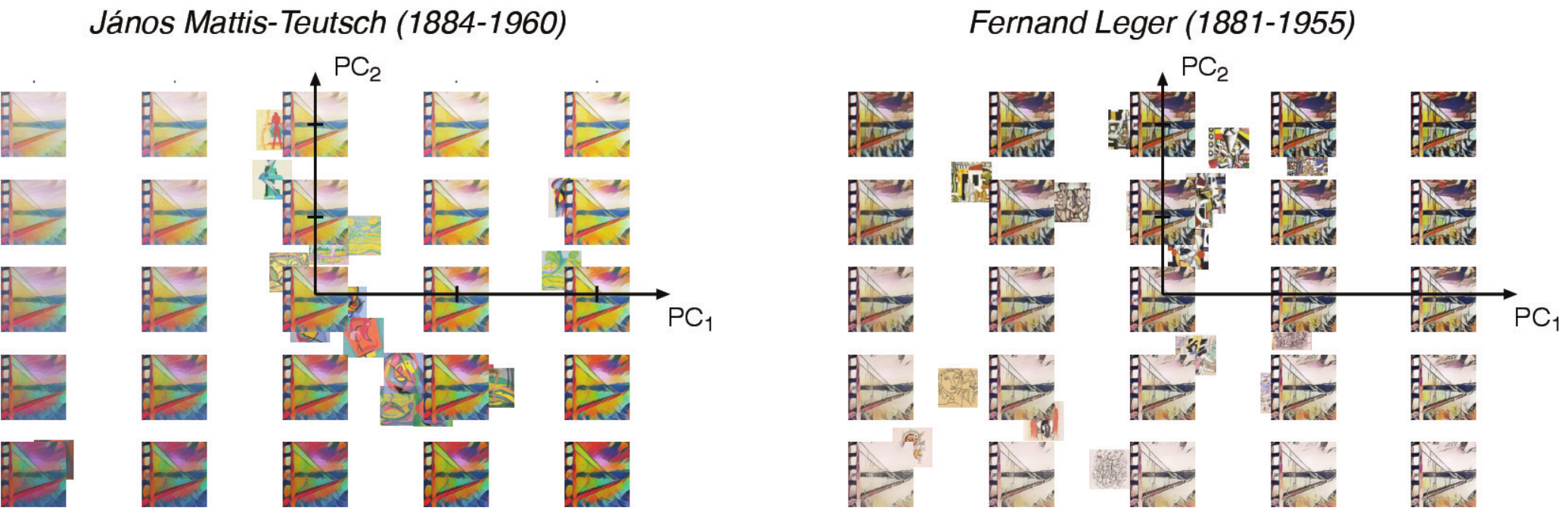}
\end{center}
\caption{Exploring the artistic range of an artist using the embedding representation.
Calculated two-dimensional principal components for a given artist
and plotted paintings from artist in this space. The principal component space
is graphically depicted by the artistic stylizations rendered on a photograph
of the Golden Gate Bridge. The center rendering is the mean and each axis
spans $\pm 4$ standard deviations in along each axis. Each axis tick mark
indicates 2 standard deviations.
Left: Paintings and principal components of Janos Mattis-Teutsch (1884-1960).
Right: Paintings and principal components of Fernand Leger (1881-1955).
Please zoom in on electronic version for details.}
\label{fig:exploration}
\end{figure*}

To explore the embedding structure further, we examined whether we can generate reasonable stylizations by varying local style changes for a specific painting style. 
In detail, we calculate the average embedding of the paintings from a specific artist and vary the embedding vector along along the two principal components of the cluster.
Figure~\ref{fig:exploration} shows stylizations from these embedding variations in a 5x5 grid, together with actual paintings of the artist whose embeddings are nearby the grid.
The stylizations from the grid captures two axis of style variations and correspond well to the neighboring embeddings of actual paintings. 
The results suggest that the model might capture a local manifold from an individual artist or painting style.


\begin{figure*}
\begin{center}
\includegraphics[width=0.9\linewidth]{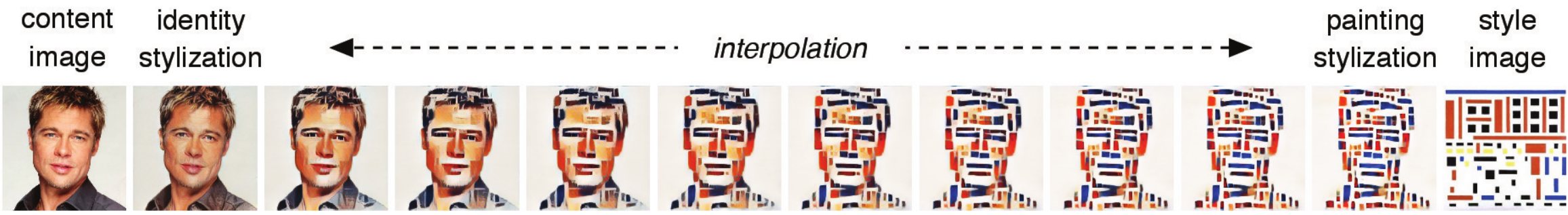}
\end{center}
\caption{Linear interpolation between identity transformation
and unobserved painting. Note that the identity transformation
is performed by feeding in the content image as the style image.}
\label{fig:interpolation}
\end{figure*}

Although we trained the style prediction network on painting
images, we find that embedding representation is extremely flexible. In particular,
supplying the network with a content image (i.e. photograph)
produces an embedding that acts as the identity tranformation.
Figure \ref{fig:interpolation} highlights the identity transformation on
a given content image. Importantly, we can now interpolate between the identity
stylization and arbitrary (in this case, unobserved) painting in order to
effectively dial in the weight of the painting style.

\section{Conclusions}
\label{sec:conclusions}

We have presented a new method for performing fast, arbitrary artistic style
transfer
on images. This model is trained at a large scale and generalizes to perform
stylizations based on paintings never previously observed. Importantly, we
demonstrate that increasing the corpus of trained painting style
confers the system the ability to generalize to
unobserved painting styles. We demonstrate that the ability to generalize
is largely predictable based on the proximity of the unobserved style to styles
trained on by the model.

We find that the model architecture provides a low dimensional
embedding space of normalization constants that captures many semantic properties
of paintings. We explore this space by demonstrating a low dimensional
space that captures the artistic range and vocabulary of a given artist.
In addition, we introduce a new form of interpolation that allows a user to arbitrarily
to dial in the strength of an artistic stylization.

This work offers several directions for future exploration. In particular,
we observe that the embedding representation for paintings only captures a portion
of the semantic information available for a painting. One might leverage
metadata of paintings in order to refine the embedding representation through
a secondary embedding loss \cite{frome2013devise, norouzi2013zero}.
Another direction is to improve the visual quality of the artistic stylization
through complementary methods that preserve the color of the original
photograph or restrict the stylization to a spatial region of the image \cite{gatys2}.
In addition, in a real
time video, one could train the network to enforce temporal consistency
between frames by appending additional loss functions \cite{ruder2016artistic}.

Aside from providing another tool for manipulating photographs, artistic style
transfer offers several applications and opportunities. Much
work in robotics has focused on training models in simulated environments with
the goal of applying this training in real world environments. Improved
stylization techniques may provide an opportunity to improve generalization
to real-world domains where data is limited \cite{bousmalis2016}.
Furthermore, by building models of paintings with low dimensional representation for painting style, we hope these representation might offer some insights into the complex statistical dependencies in paintings if not images in general to improve our understanding of the structure of natural image statistics.

\cutsectionup
\section{Acknowledgments}
\label{sec:conclusions}
\cutsectiondown

We wish to thank Douglas Eck, Fred Bertsch for valuable discussions and support;
Xin Pan for technical assistance and advice;
the larger Google Brain team for valuable discussion and support.


{\small
\bibliographystyle{abbrv}
\bibliography{paper_arxiv}

\begin{thebibliography}{10}

\bibitem{bousmalis2016}
K.~Bousmalis, G.~Trigeorgis, N.~Silberman, D.~Krishnan, and D.~Erhan.
\newblock Domain separation networks.
\newblock In D.~D. Lee, M.~Sugiyama, U.~V. Luxburg, I.~Guyon, and R.~Garnett,
  editors, {\em Advances in Neural Information Processing Systems 29}, pages
  343--351. Curran Associates, Inc., 2016.

\bibitem{cimpoi14describing}
M.~Cimpoi, S.~Maji, I.~Kokkinos, S.~Mohamed, , and A.~Vedaldi.
\newblock Describing textures in the wild.
\newblock In {\em Proceedings of the {IEEE} Conf. on Computer Vision and
  Pattern Recognition ({CVPR})}, 2014.

\bibitem{dumoulin2016}
V.~Dumoulin, J.~Shlens, and M.~Kudlur.
\newblock A learned representation for artistic style.
\newblock {\em International Conference of Learned Representations (ICLR)},
  2016.

\bibitem{efros2001image}
A.~A. Efros and W.~T. Freeman.
\newblock Image quilting for texture synthesis and transfer.
\newblock In {\em Proceedings of the 28th annual conference on Computer
  graphics and interactive techniques}, pages 341--346. ACM, 2001.

\bibitem{efros1999texture}
A.~A. Efros and T.~K. Leung.
\newblock Texture synthesis by non-parametric sampling.
\newblock In {\em Computer Vision, 1999. The Proceedings of the Seventh IEEE
  International Conference on}, volume~2, pages 1033--1038. IEEE, 1999.

\bibitem{freeman2011metamers}
J.~Freeman and E.~P. Simoncelli.
\newblock Metamers of the ventral stream.
\newblock {\em Nature neuroscience}, 14(9):1195--1201, 2011.

\bibitem{frome2013devise}
A.~Frome, G.~S. Corrado, J.~Shlens, S.~Bengio, J.~Dean, T.~Mikolov, et~al.
\newblock Devise: A deep visual-semantic embedding model.
\newblock In {\em Advances in neural information processing systems}, pages
  2121--2129, 2013.

\bibitem{gatys2015texture}
L.~Gatys, A.~S. Ecker, and M.~Bethge.
\newblock Texture synthesis using convolutional neural networks.
\newblock In {\em Advances in Neural Information Processing Systems}, pages
  262--270, 2015.

\bibitem{gatys2015neural}
L.~A. Gatys, A.~S. Ecker, and M.~Bethge.
\newblock A neural algorithm of artistic style.
\newblock {\em arXiv preprint arXiv:1508.06576}, 2015.

\bibitem{gatys2}
L.~A. Gatys, A.~S. Ecker, M.~Bethge, A.~Hertzmann, and E.~Shechtman.
\newblock Controlling perceptual factors in neural style transfer.
\newblock {\em CoRR}, abs/1611.07865, 2016.

\bibitem{hertzmann2001image}
A.~Hertzmann, C.~E. Jacobs, N.~Oliver, B.~Curless, and D.~H. Salesin.
\newblock Image analogies.
\newblock In {\em Proceedings of the 28th annual conference on Computer
  graphics and interactive techniques}, pages 327--340. ACM, 2001.

\bibitem{huang2017}
X.~Huang and S.~Belongie.
\newblock Arbitrary style transfer in real-time with adaptive instance
  normalization.
\newblock {\em arXiv preprint arXiv:1703.06868}, 2017.

\bibitem{irving1969}
C.~Irving.
\newblock {\em Fake: the story of Elmyr de Hory: the greatest art forger of our
  time.}
\newblock McGraw-Hill, 1969.

\bibitem{johnson2016perceptual}
J.~Johnson, A.~Alahi, and L.~Fei-Fei.
\newblock Perceptual losses for real-time style transfer and super-resolution.
\newblock {\em arXiv preprint arXiv:1603.08155}, 2016.

\bibitem{julesz1962}
B.~Julesz.
\newblock Visual pattern discrimination.
\newblock {\em IRE Trans. Info Theory}, 8:84--92, 1962.

\bibitem{liwand2016}
C.~Li and M.~Wand.
\newblock Precomputed real-time texture synthesis with markovian generative
  adversarial networks.
\newblock {\em ECCV}, 2016.

\bibitem{liang2001real}
L.~Liang, C.~Liu, Y.-Q. Xu, B.~Guo, and H.-Y. Shum.
\newblock Real-time texture synthesis by patch-based sampling.
\newblock {\em ACM Transactions on Graphics (ToG)}, 20(3):127--150, 2001.

\bibitem{maaten2008visualizing}
L.~v.~d. Maaten and G.~Hinton.
\newblock Visualizing data using t-sne.
\newblock {\em Journal of Machine Learning Research}, 9(Nov):2579--2605, 2008.

\bibitem{inceptionism}
A.~Mordvintsev, C.~Olah, and M.~Tyka.
\newblock Inceptionism: Going deeper into neural networks, June 2015.

\bibitem{norouzi2013zero}
M.~Norouzi, T.~Mikolov, S.~Bengio, Y.~Singer, J.~Shlens, A.~Frome, G.~S.
  Corrado, and J.~Dean.
\newblock Zero-shot learning by convex combination of semantic embeddings.
\newblock {\em arXiv preprint arXiv:1312.5650}, 2013.

\bibitem{portilla1999}
J.~Portilla and E.~Simoncelli.
\newblock A parametric texture model based on joint statistics of complex
  wavelet coefficients.
\newblock {\em International Journal of Computer Vision}, 40:49--71, 1999.

\bibitem{ruder2016artistic}
M.~Ruder, A.~Dosovitskiy, and T.~Brox.
\newblock Artistic style transfer for videos.
\newblock In {\em German Conference on Pattern Recognition}, pages 26--36.
  Springer, 2016.

\bibitem{simonyan2014very}
K.~Simonyan and A.~Zisserman.
\newblock Very deep convolutional networks for large-scale image recognition.
\newblock {\em arXiv preprint arXiv:1409.1556}, 2014.

\bibitem{szegedy2015}
C.~Szegedy, V.~Vanhoucke, S.~Ioffe, J.~Shlens, and Z.~Wojna.
\newblock Rethinking the inception architecture for computer vision.
\newblock {\em IEEE Computer Vision and Pattern Recognition (CVPR)}, 2015.

\bibitem{ulyanov2016texture}
D.~Ulyanov, V.~Lebedev, A.~Vedaldi, and V.~Lempitsky.
\newblock Texture networks: Feed-forward synthesis of textures and stylized
  images.
\newblock {\em arXiv preprint arXiv:1603.03417}, 2016.

\bibitem{ulyanov2016instance}
D.~Ulyanov, A.~Vedaldi, and V.~Lempitsky.
\newblock Instance normalization: The missing ingredient for fast stylization.
\newblock {\em arXiv preprint arXiv:1607.08022}, 2016.

\bibitem{wei2000fast}
L.-Y. Wei and M.~Levoy.
\newblock Fast texture synthesis using tree-structured vector quantization.
\newblock In {\em Proceedings of the 27th annual conference on Computer
  graphics and interactive techniques}, pages 479--488. ACM
  Press/Addison-Wesley Publishing Co., 2000.

\bibitem{zeiler2014visualizing}
M.~D. Zeiler and R.~Fergus.
\newblock Visualizing and understanding convolutional networks.
\newblock In {\em European Conference on Computer Vision}, pages 818--833.
  Springer, 2014.

\end{thebibliography}
}

\newpage
\appendix
\setcounter{figure}{0}
\setcounter{table}{0}
\section{Supplementary}

\newcommand{\ourres}{figures_supplementary/PBN1024test/ours/}
\newcommand{\adaInres}{figures_supplementary/PBN1024test/AdaIN/}
\newcommand{\gatysres}{figures_supplementary/PBN1024test/Gatys/}
\newcommand{\deepartio}{figures_supplementary/PBN1024test/deepartio/}

\subsection{Variety of stylizations in trained model.}
 The figure below shows mosaic of stylizations across 2592 paintings using model trained on PBN dataset.

\begin{figure}[H]
\centering
\includegraphics[width=1\linewidth]{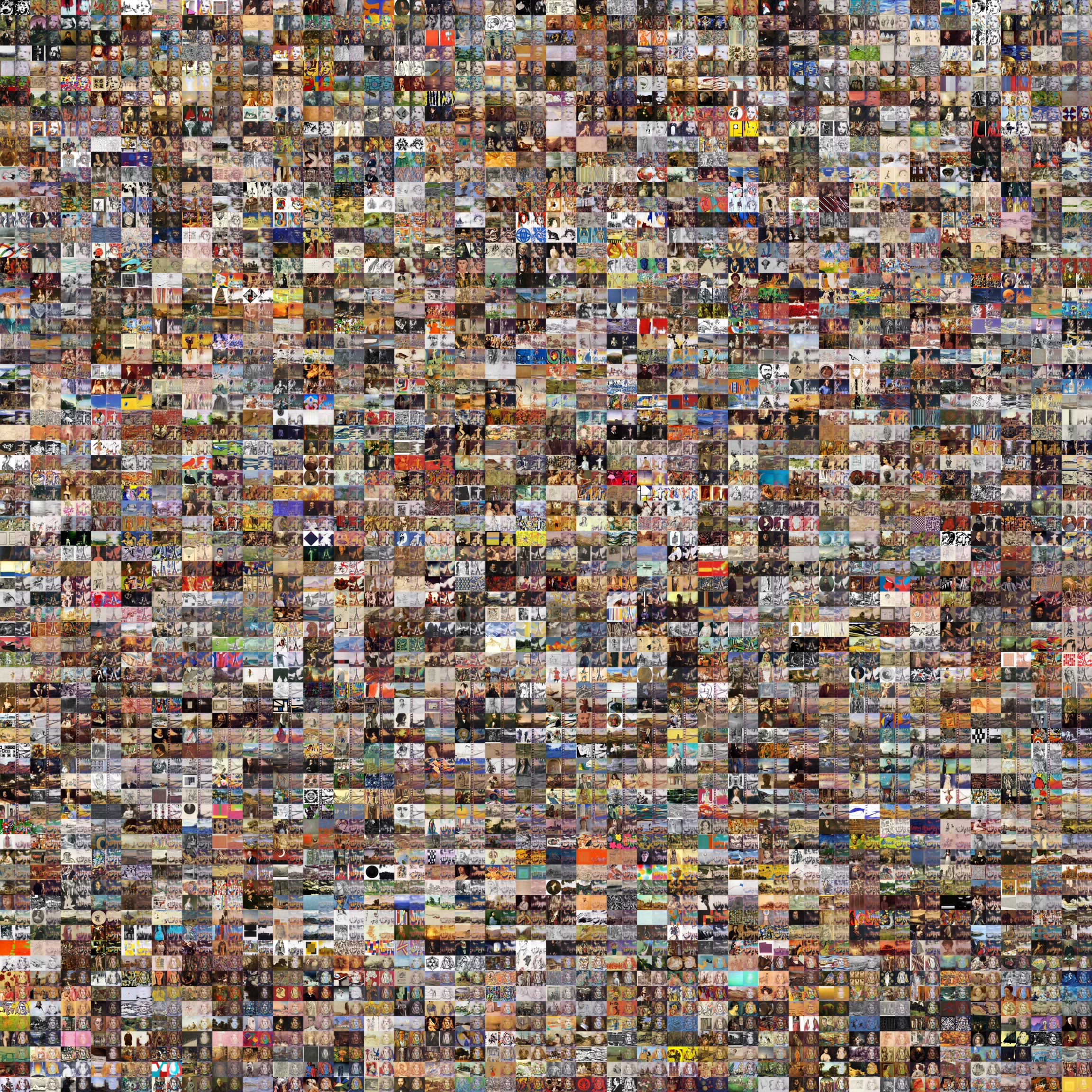}\\
\caption{Mosaic of stylizations across 2592 paintings using model trained on PBN
dataset. Left-hand side shows painting and right-hand side shows
stylization across an assortment of 8 content images.
Please zoom in on a digital copy to examine the details of the individual paintings.}
\end{figure}
\newpage

\subsection{Structure of a low-dimensional representation of the embedding space}
\begin{figure}[H]
\centering
\begin{tabular}{c}
\includegraphics[width=0.55\linewidth]{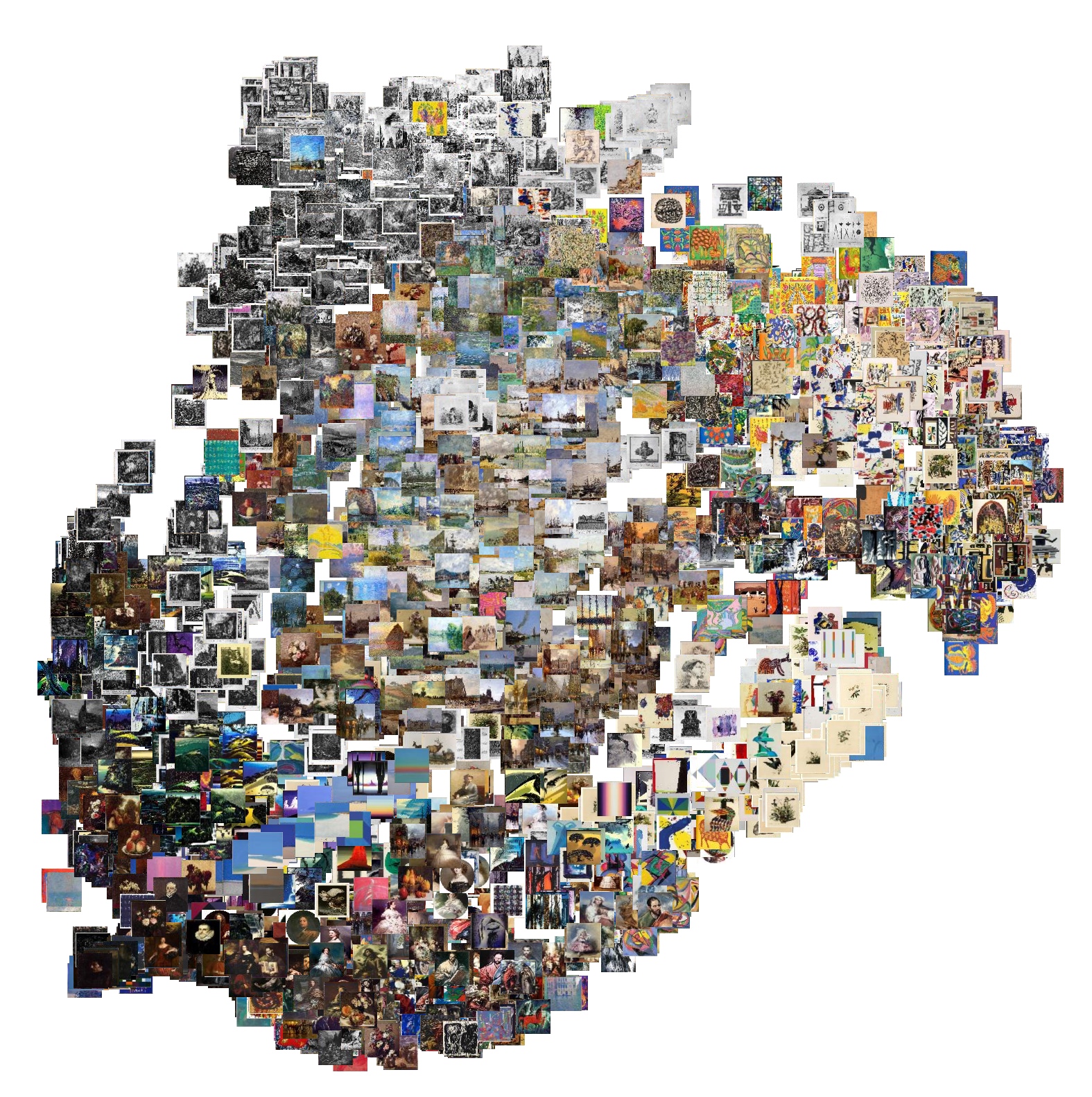}\\
\includegraphics[width=0.55\linewidth]{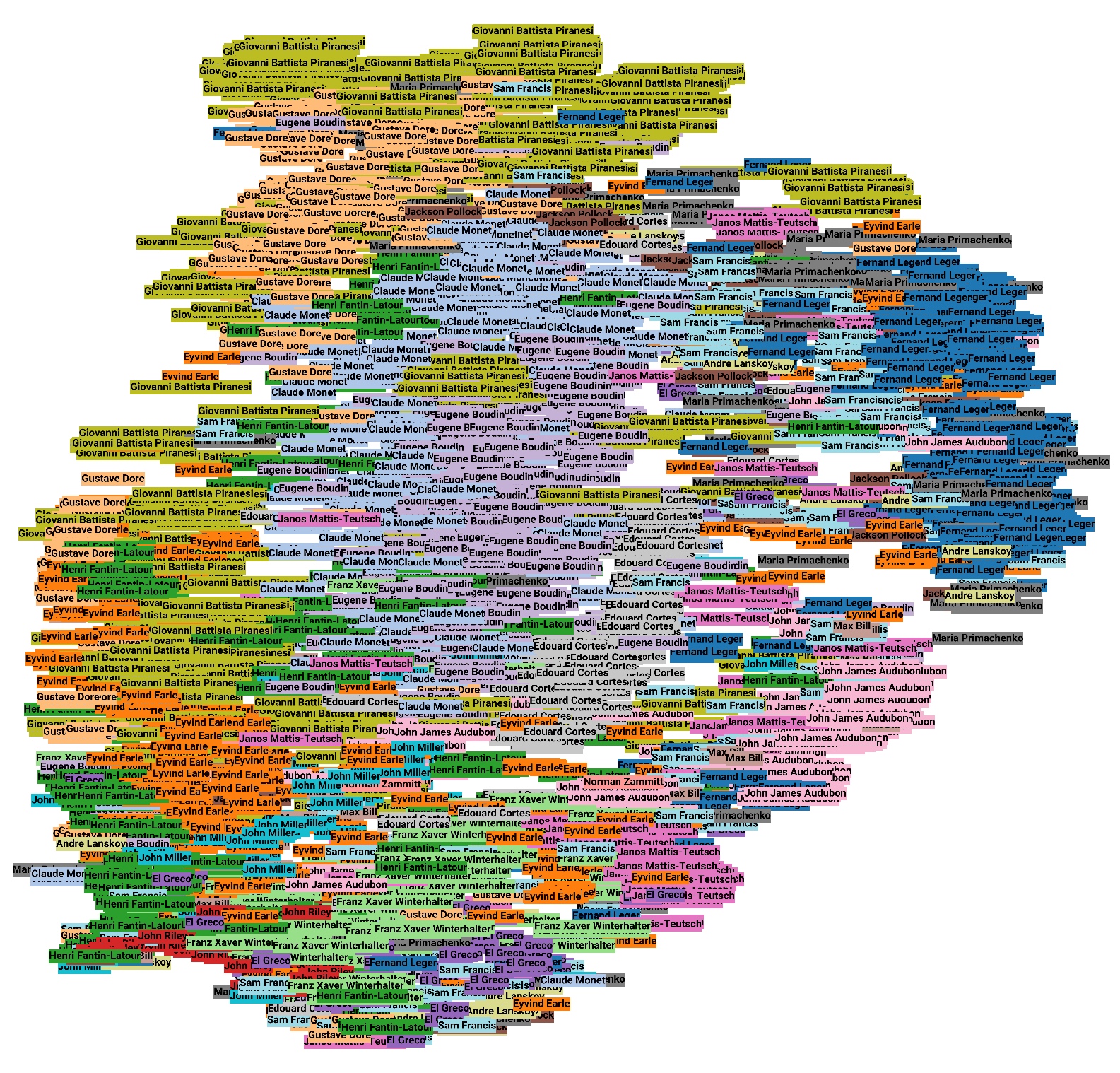}\\
\end{tabular}
\caption{Top: Expanded view of t-SNE representation for low-dimensional style
embedding space for PBN dataset of paintings across 20 painters. Bottom:
Same as above but replacing the painting with metadata indicating the artist.
Please zoom in on a digital copy to examine the details of the individual paintings.}
\label{fig:tsneA}
\end{figure}

\begin{figure}[H]
\centering
\begin{tabular}{c}
\includegraphics[width=0.61\linewidth]{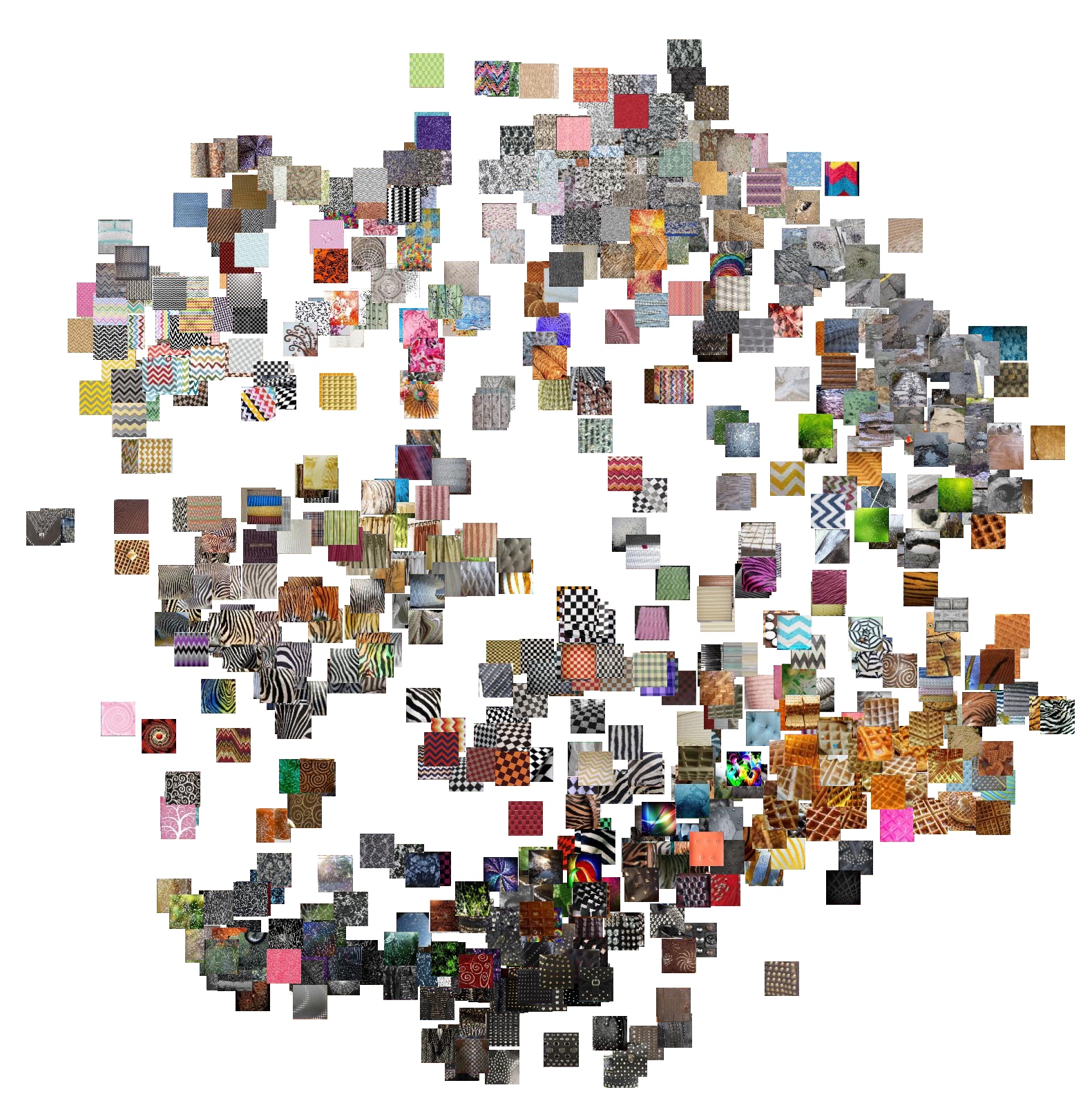}\\
\includegraphics[width=0.61\linewidth]{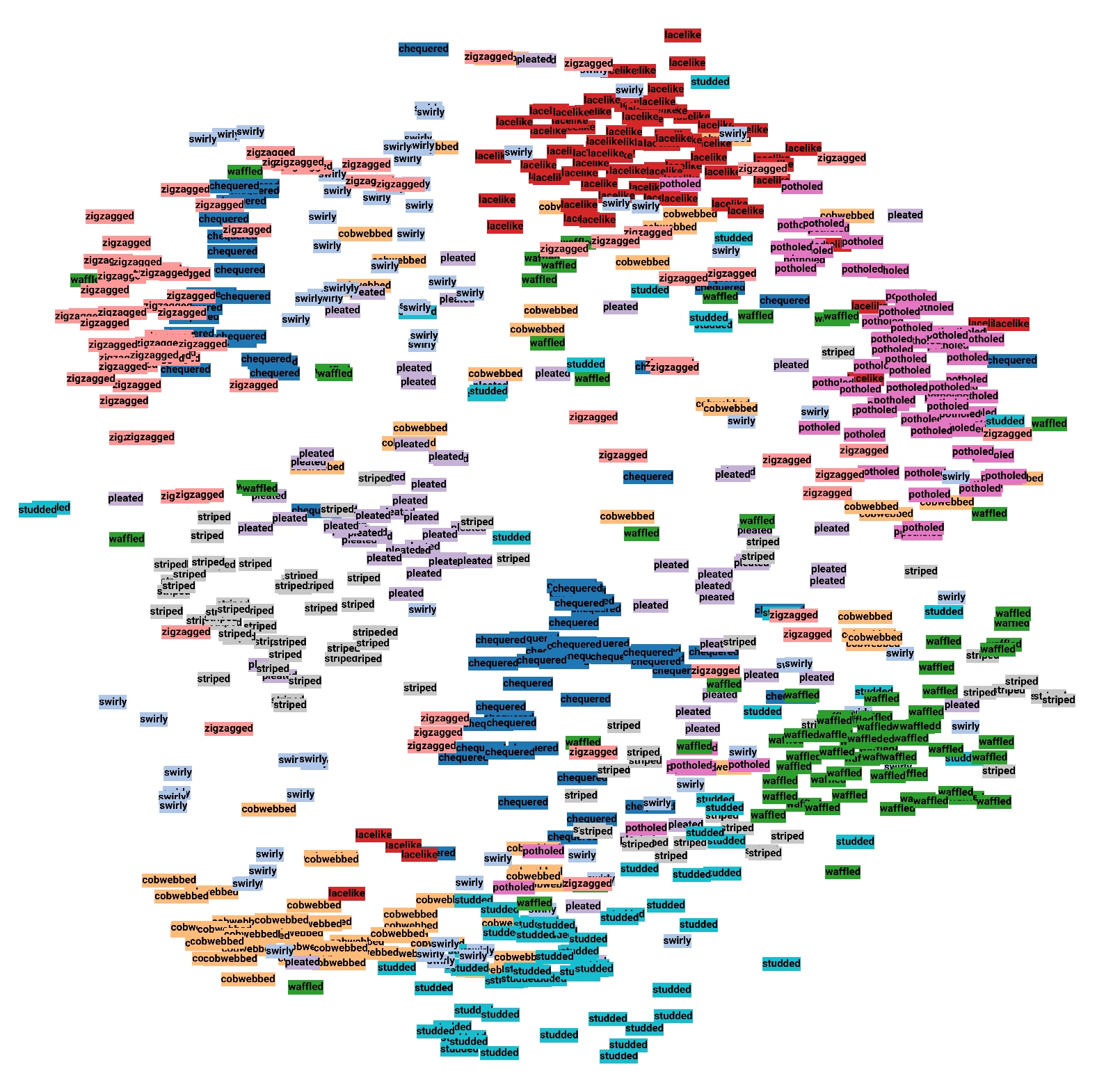}\\
\end{tabular}
\caption{Top: Expanded view of t-SNE representation for low-dimensional style
embedding space for DTD visual textures dataset across 10 categories. 
Bottom: Same as above but replacing the visual texture with metadata describing the texture.
Please zoom in on a digital copy to examine the details of the individual textures.}
\label{fig:tsneB}
\end{figure}

In Figure \ref{fig:tsneA} and Figure \ref{fig:tsneB}, we provide
high-resolution version of the t-SNE embeddings learned from the painting
and texture datasets, respectively.

\newpage
\subsection{Qualitative comparison with other methods}
In this section, we provide comparison of our method trained on PBN 
training images, AdaIN[12], Gatys[9] (using the same set of parameters for all the images)
and deepart (\texttt{https://deepart.io})
across many style and content images.

\vspace{1cm}

\newcommand{\insertnames}[1]{%
	\scriptsize
	\ifthenelse{\equal{#1}{avril}}%
	{%
	\begin{tabularx}{0.88\linewidth}{*{6}{>{\centering\arraybackslash}X}}
		\textbf{content image} & \textbf{style image} & \textbf{Ours} & \textbf{AdaIN} & \textbf{Gatys et al} & \textbf{deepart.io}\\
	\end{tabularx}\\
	}%
	{}%
}

\newcommand{\clearthepage}[1]{%
	\ifthenelse{\equal{#1}{karya}}%
	{%
	\clearpage
	}%
	{}%
}

\foreach \stylename/\contentname in{
	9121/avril,101224/azadi,100188/christ,100733/eiffel,1564/golden_gate,7952/gondol_in_venizia,23808/karya,
	65422/avril,62959/azadi,70435/beach,40705/christ,41932/eiffel,42573/golden_gate,44693/gondol_in_venizia,50285/karya,
	50343/avril,52606/azadi,55630/beach,55730/christ,55822/eiffel,56096/golden_gate,56686/gondol_in_venizia,58186/karya,
	102189/avril,73553/azadi,1226/beach,83004/christ,14739/eiffel,97467/golden_gate,95631/gondol_in_venizia,70718/karya,
	11682/avril,17242/azadi,21284/beach,3545/christ,75491/eiffel,10166/golden_gate,10892/gondol_in_venizia,497/karya}
{
\begin{center}
    \insertnames{\contentname}
    \includegraphics[width=0.14\linewidth]{\ourres/\contentname .png}
    \includegraphics[width=0.14\linewidth]{\ourres/\stylename .png}
    \includegraphics[width=0.14\linewidth]{\ourres/\contentname _stylized_\stylename .png}
    \includegraphics[width=0.14\linewidth]{\adaInres/\contentname _stylized_\stylename .jpg}
    \includegraphics[width=0.14\linewidth]{\gatysres/{\contentname .png__stylized_\stylename .png}.png}
    \color{white}
    \includegraphics[width=0.14\linewidth]{\deepartio/\contentname _stylized_\stylename .jpg}
    \color{black}
    \clearthepage{\contentname}
\end{center}
}

\subsection{Generalization of the model across training datasets}
\vspace{1cm}
In this section, we provide additional experiments demonstrating the 
degree of generalization of the model trained from two different training
datasets (i.e., painting vs texture). More specifically, we investigated 
the following questions. (1) How well does the model generalize from the training
data to the test data within the same domain? (2) How similar is the stylization performance when 
learned from one domain (e.g., painting images) to the stylization performance when learned 
from a different domain (e.g., texture images)? 
 
\begin{figure}[h]
\centering
\begin{tabular}{c}
	\includegraphics[width=0.45\linewidth]{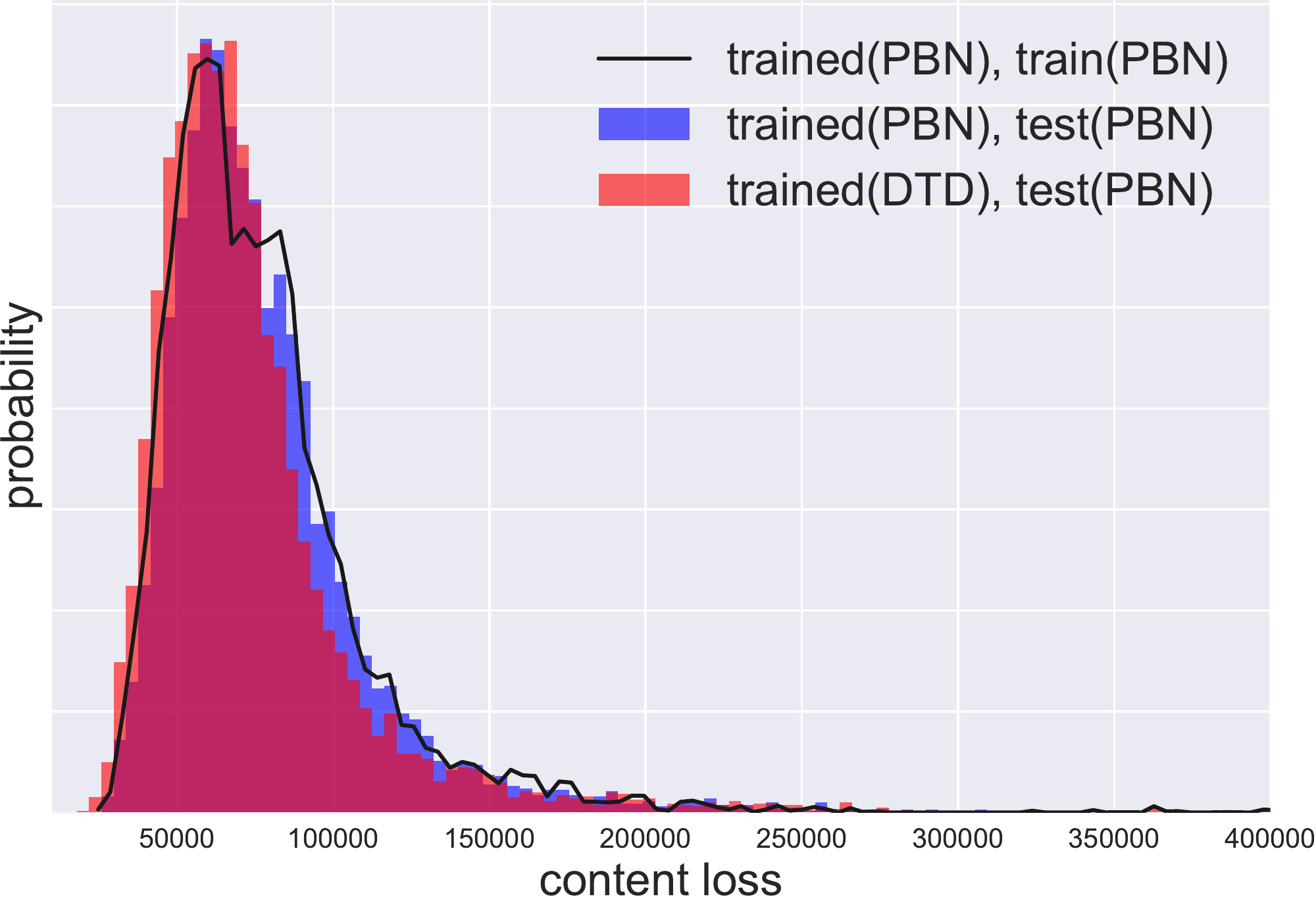}
	\includegraphics[width=0.45\linewidth]{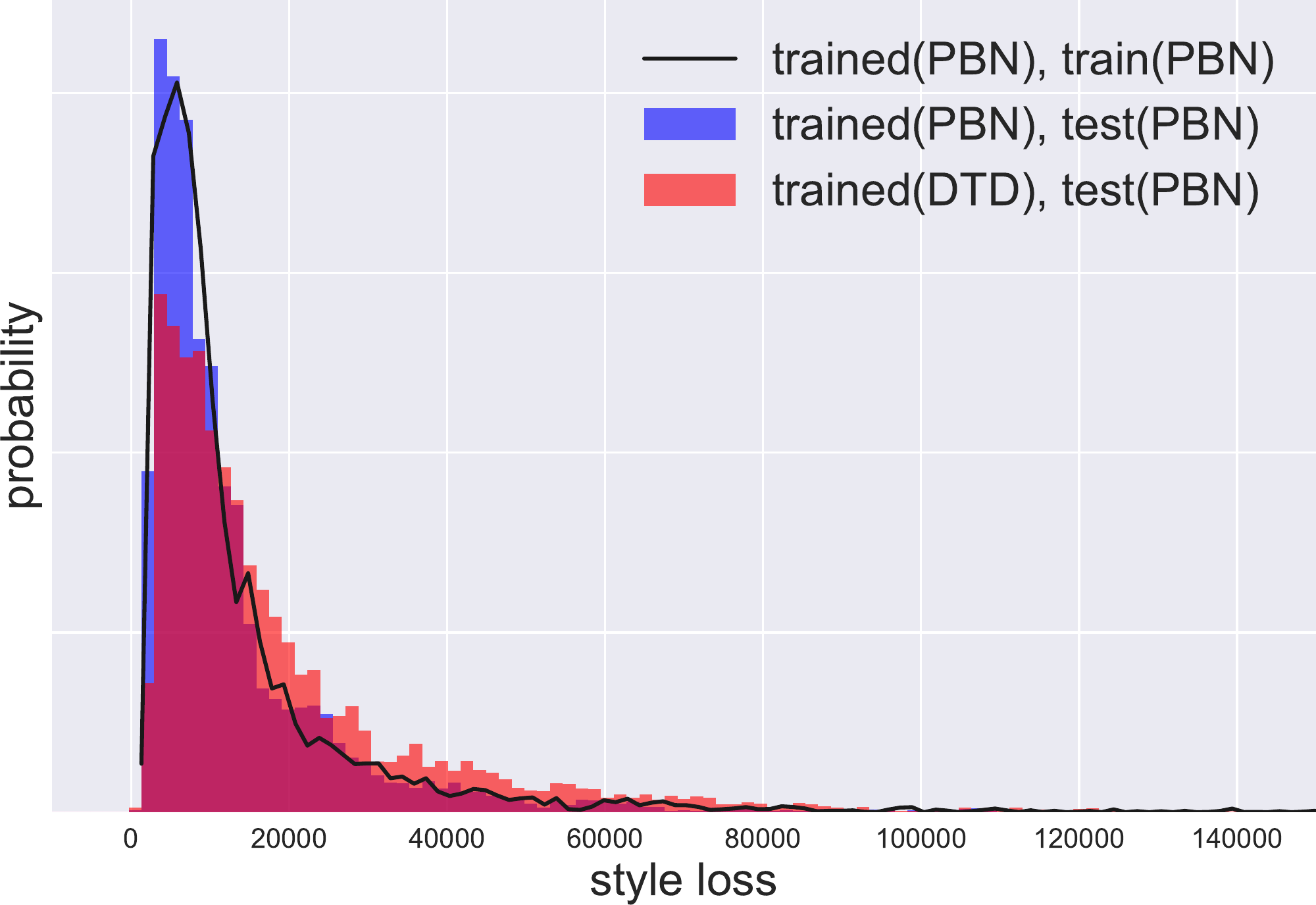}\\
	(a) Content and style losses over PBN dataset\\
	\vspace{0.2cm}
	\includegraphics[width=0.45\linewidth]{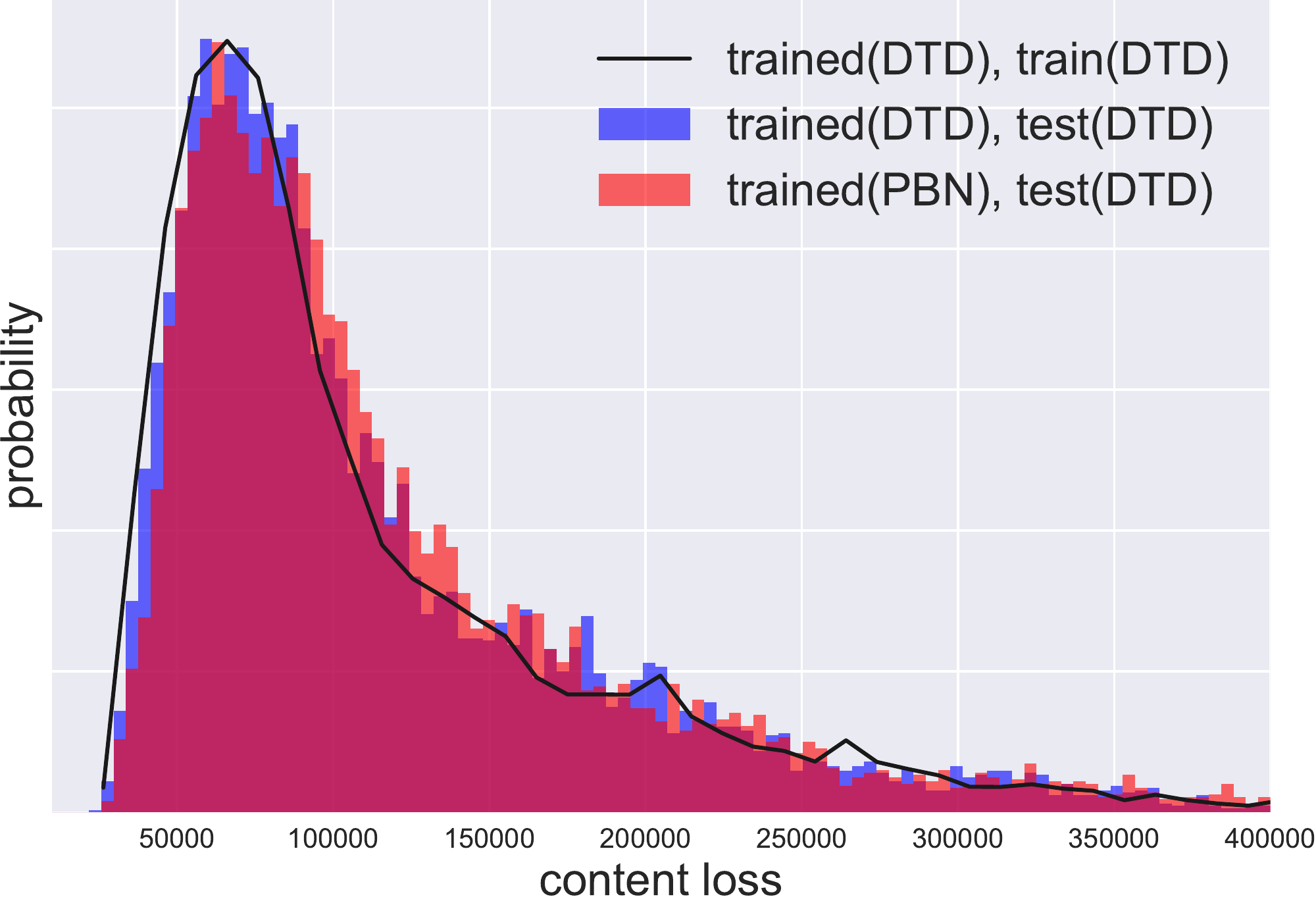}
	\includegraphics[width=0.45\linewidth]{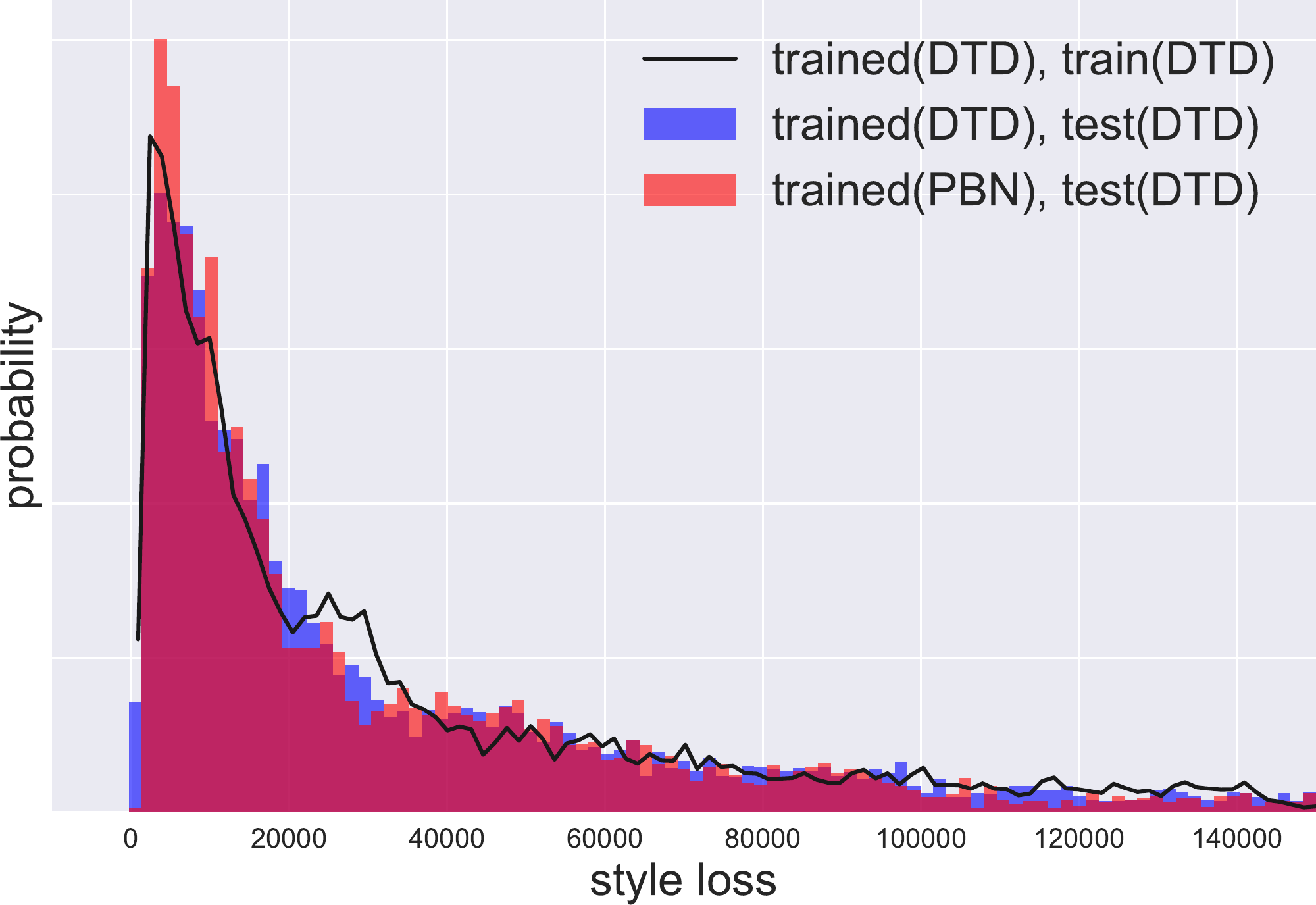}\\
	(b) Content and style losses over DTD dataset\\
\end{tabular}
\vspace{0.2cm}
\caption{Measuring the ability of the model to generalize across datasets.
Black curves represent the distribution of style and content losses for the 
training images in the DTD and PBN datasets, respectively. Blue histograms 
represent the distributions for testing images for a model trained on the 
training images of the same dataset. Red histograms represent the distributions
for testing images for a model trained on a training set of a different dataset.
The two models trained on DTD and PBN are trained with the exact same parameters
except the content weight for the model trained on DTD dataset is larger.}
\label{fig:cross_datasets}
\end{figure}

In response to the first question, we showed in Section 3.2 that the 
distribution of the style loss for the training style images and test style
images closely matches  (Figure 3 of the main paper), suggesting that the model
generalizes well within the same domain. For the second question, we measured 
the distribution of style loss and content loss when a model is trained and 
tested on different datasets (i.e., training from painting images and testing
on texture images and vice versa). Figure~\ref{fig:cross_datasets} shows the 
summary histograms.
For this experiment we calculate the distribution of
style and content losses across 8 photographs for 1024 unobserved painting
styles. (Statistics in section 3.2 are calculated acrross 2 photographs).
Surprisingly, we found that the distribution of style loss
is very similar regardless of which dataset of style images the model was 
trained from. For example, Figure~\ref{fig:cross_datasets}(a) shows that the 
model trained from the painting dataset and the model trained from the texture
dataset produce similar style loss and content loss distributions when evaluated
on the same test painting images. Similar result is shown when using the
texture dataset for evaluation, as shown in Figure~\ref{fig:cross_datasets}(b). The summary
statistics in Table \ref{table:summary_cross_dtd} and
Table \ref{table:summary_cross_pbn} also support this conclusion. 
Furthermore, the stylizations of the two different models are perceptually
similar, as shown in the figure panels below. These results suggest that 
when we train from a sufficiently large corpus of style images (which covers
a rich variety of color and texture), the learned model might be able to 
generalize even to unseen types of images.
\vspace{2cm}

\newcommand{\ourrespbn}{figures_supplementary/PBN1024test/ours/}
\newcommand{\ourdtdrespbn}{figures_supplementary/PBN1024test/ours_trained_dtd_10M/}
\newcommand{\ourresdtd}{figures_supplementary/DTD1024test/ours/}
\newcommand{\ourdtdresdtd}{figures_supplementary/DTD1024test/ours_trained_dtd_10M/}
\renewcommand{\adaInres}{figures_supplementary/DTD1024test/AdaIN/}
\renewcommand{\gatysres}{figures_supplementary/DTD1024test/Gatys/}
\newcommand{\cmpourswid}{0.12}

\newcommand{\ainsertnames}[1]{%
	\scriptsize
	\ifthenelse{\equal{#1}{avril}}%
	{%
	\begin{tabularx}{0.14\linewidth}{*{1}{>{\centering\arraybackslash}X}}
		\textbf{content image}
	\end{tabularx}
	\hspace{0.5cm}
	\begin{tabularx}{0.37\linewidth}{*{3}{>{\centering\arraybackslash}X}}
		\textbf{style image (PBN test)} & \textbf{trained on PBN} & \textbf{trained on DTD}
	\end{tabularx}
	\hspace{0.5cm}
	\begin{tabularx}{0.37\linewidth}{*{3}{>{\centering\arraybackslash}X}}
		\textbf{style image (DTD test)} & \textbf{trained on PBN} & \textbf{trained on DTD}
	\end{tabularx}\\
	}%
	{}%
}

\begin{table}[h]
\scriptsize
\centering
\begin{tabular}{|r||c|c|c||c|c|c|}
	\hline
	&\multicolumn{3}{|c||}{content loss}&\multicolumn{3}{|c|}{style loss}\\
	\hline
	&mean& median& std &mean& median& std\\
	\hline
	trained(DTD), train(DTD)& 117230 & 86840 & 90640 & 56840& 22230& 152870\\
	trained(DTD), test(DTD)& 117710 & 90380 & 79430 & 56420& 19780& 133350\\
	trained(PBN), test(DTD)& 123270 & 96110 & 83760 & 75940& 18320& 285110\\
	\hline
\end{tabular}
\caption{Summary statistics of content and style loss over DTD dataset.}
\label{table:summary_cross_dtd}
\end{table}

\begin{table}[h]
\scriptsize
\centering
\begin{tabular}{|r||c|c|c||c|c|c|}
	\hline
	&\multicolumn{3}{|c||}{content loss}&\multicolumn{3}{|c|}{style loss}\\
	\hline
	&mean& median& std &mean& median& std\\
	\hline
	trained(PBN), train(PBN)& 77750 &70650&34290 & 14000 &8420 & 19000\\
	trained(PBN), test(PBN)& 79070 &71310 &36700 & 14730 &8593 & 30750\\
	trained(DTD), test(PBN)& 73620 &66030 &35690 & 24360 &12810 & 57350\\
	\hline
\end{tabular}
\caption{Summary statistics of content and style loss over PBN dataset.}
\label{table:summary_cross_pbn}
\end{table}
\vspace{3cm}

\newpage
\subsubsection{Qualitative comparisons of generalization across datasets.}
Qualitative comparisons of our method trained on the PBN training images and our method trained on the DTD 
training images on some style images from PBN test data and DTD test data are included in the follwoing pages.
\vspace{2cm}
\foreach \pstylename/\dstylename/\contentname in{
	10892/braided_0078/avril,
	56096/banded_0152/avril,98508/chequered_0179/azadi,83004/cobwebbed_0157/beach,41932/cracked_0162/christ,90218/zigzagged_0129/eiffel,21284/studded_0119/golden_gate,70435/sprinkled_0059/gondol_in_venizia,101224/dotted_0154/karya,
	97467/fibrous_0116/avril,55730/lacelike_0098/azadi,52606/frilly_0079/beach,14739/grid_0064/christ,20021/grid_0105/eiffel,23457/honeycombed_0133/golden_gate,26130/striped_0043/gondol_in_venizia,80687/spiralled_0070/karya}
{
\begin{center}
    \ainsertnames{\contentname}
    \includegraphics[width=\cmpourswid\linewidth]{\ourrespbn/\contentname .png}
    \hspace{0.5cm}
    \includegraphics[width=\cmpourswid\linewidth]{\ourrespbn/\pstylename .png}
    \includegraphics[width=\cmpourswid\linewidth]{\ourrespbn/\contentname _stylized_\pstylename .png}
    \includegraphics[width=\cmpourswid\linewidth]{\ourdtdrespbn/\contentname _stylized_\pstylename .png}
    \hspace{0.5cm}
    \includegraphics[width=\cmpourswid\linewidth]{\ourresdtd/\dstylename .png}
    \includegraphics[width=\cmpourswid\linewidth]{\ourresdtd/\contentname _stylized_\dstylename .png}
    \includegraphics[width=\cmpourswid\linewidth]{\ourdtdresdtd/\contentname _stylized_\dstylename .png}
    \vspace{0.5cm}
    \clearthepage{\contentname}
\end{center}
}

\subsection{Expanded analysis on the generalization to unobserved paintings}

In Section 3.2 of the main paper (Figure 4), we discussed how the 
generalization ability of the model (measured in terms of style loss)
is related to the proximity of training examples. In this section, we 
provide more detailed analysis with qualitative results. 
 
\begin{figure}[H]
\centering
\begin{tabular}{c}
\includegraphics[width=0.6\linewidth]{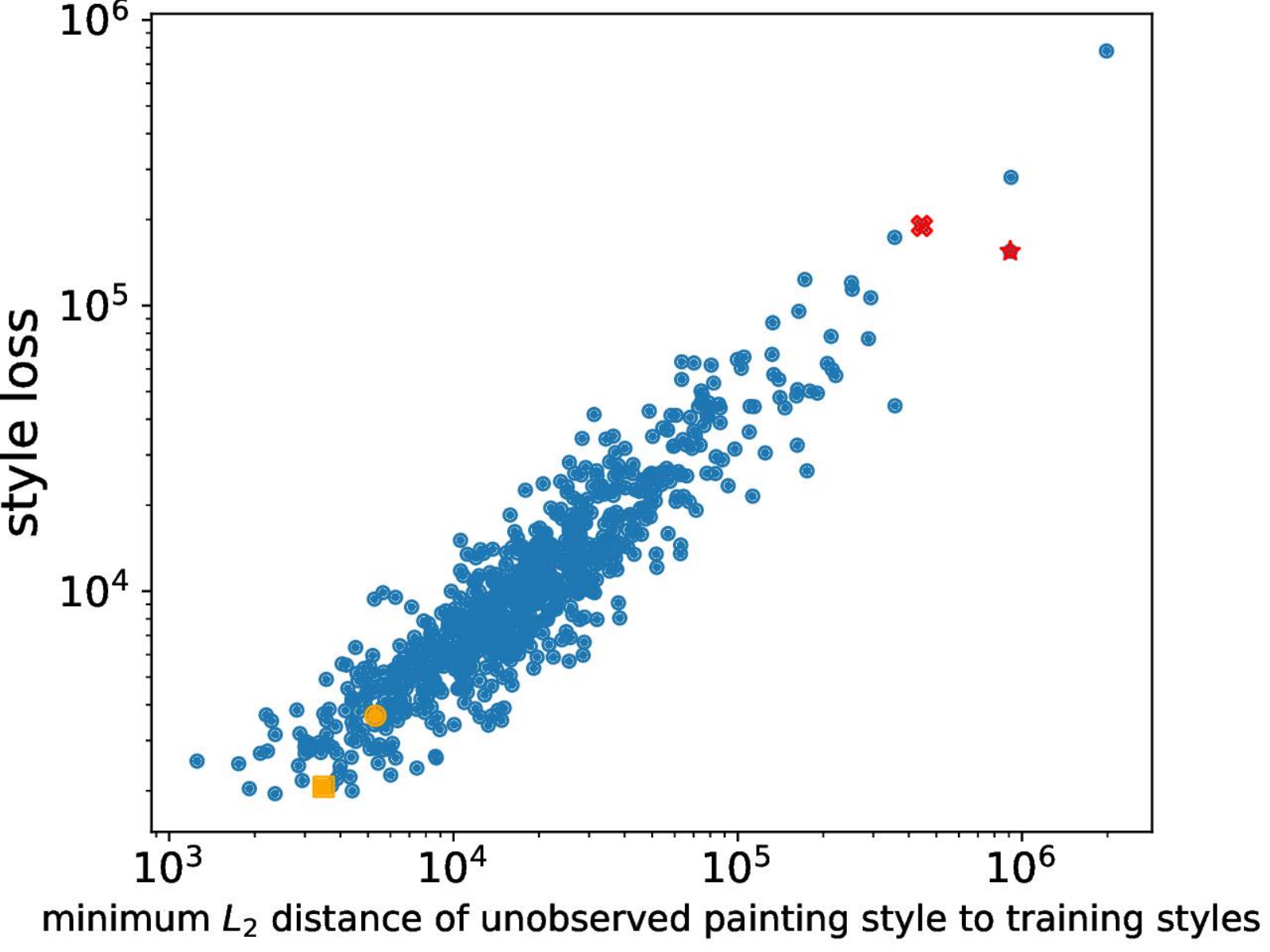}\\
\end{tabular}
\caption{Style loss vs. proximity to training examples}
\label{fig:generalization_unobserved}
\end{figure}

For better visualization, we show the scatter plot in log-log scale in
Figure~\ref{fig:generalization_unobserved}. Each point in the scatter 
corresponds to a different test style image. As described in the main 
paper, we find a clear correlation between the style loss on unobserved
paintings and the minimum L2 distance between the Gram matrix of unobserved
painting and the set of all Gram matrices in the training dataset of paintings.
This suggests that the generalization performance (as measured by the style loss)
is positively correlated to the proximity of the test style image to the corpus
of training style images. 
 
To gain further insights, we provide example stylization results in
Figure~\ref{fig:generalization_examples} for two groups. In other words,
the yellow points represent ``easy examples'' where the corresponding test style 
image is similar to the nearby training style images in the Gram matrix space 
and the resulting style loss tends to be relatively small. Conversely, the red
points represent ``difficult examples'' where the corresponding test style image
is relatively far away from the training style images in the Gram matrix space
and the style loss tends to be relatively large. For each case, we also show 
stylizations using the nearest training style image side by side. In our 
preliminary examinations, we found that the larger style loss seem to correspond
to a slightly worse stylizations in the perceptual sense. Furthermore, the
two stylizations using the test style image and the closest training style
image seem to be more perceptually different (in terms of color and texture)
for the difficult cases (red points) than the easy cases (yellow points). We
hypothesize that the difficult cases (red points) may lie around the low-density
region in the embedding space, but we leave further analysis as future work.
\vspace{2cm}

\newcommand{\pathtoexamples}{figures_supplementary/loss_vs_prosimity/examples/}
\newcommand{\proswid}{0.12}
\begin{figure}[H]
\begin{center}
\begin{tabular}{ccccc}
	\small{content image} & \small{testing style} & \small{stylization} & \small{closest training} & \small{stylization}\\
	\includegraphics[width=\proswid\linewidth]{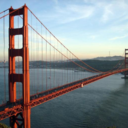}&\hspace{0.5cm}
	\includegraphics[width=\proswid\linewidth]{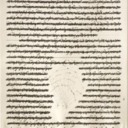}&
	\includegraphics[width=\proswid\linewidth]{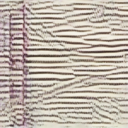}&
	\includegraphics[width=\proswid\linewidth]{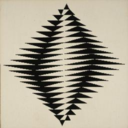}&
        \includegraphics[width=\proswid\linewidth]{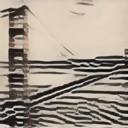}\\
	\includegraphics[width=\proswid\linewidth]{figures_supplementary/PBN1024test/ours/golden_gate.png}&\hspace{0.5cm}
	\includegraphics[width=\proswid\linewidth]{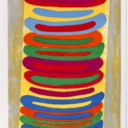}&
	\includegraphics[width=\proswid\linewidth]{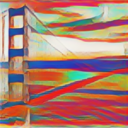}&
	\includegraphics[width=\proswid\linewidth]{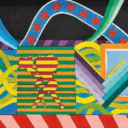}&
        \includegraphics[width=\proswid\linewidth]{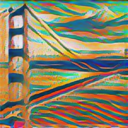}\\
	\includegraphics[width=\proswid\linewidth]{figures_supplementary/PBN1024test/ours/golden_gate.png}&\hspace{0.5cm}
	\includegraphics[width=\proswid\linewidth]{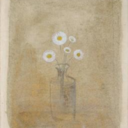}&
	\includegraphics[width=\proswid\linewidth]{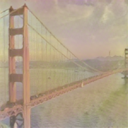}&
	\includegraphics[width=\proswid\linewidth]{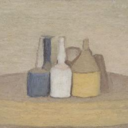}&
        \includegraphics[width=\proswid\linewidth]{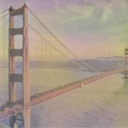}\\
	\includegraphics[width=\proswid\linewidth]{figures_supplementary/PBN1024test/ours/golden_gate.png}&\hspace{0.5cm}
	\includegraphics[width=\proswid\linewidth]{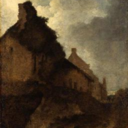}&
	\includegraphics[width=\proswid\linewidth]{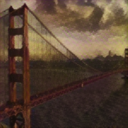}&
	\includegraphics[width=\proswid\linewidth]{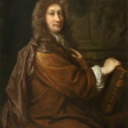}&
        \includegraphics[width=\proswid\linewidth]{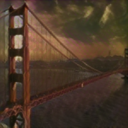}\\
\end{tabular}
\end{center}
	\caption{Some qualitive examples for the points in figure \ref{fig:generalization_unobserved}.
	First two rows are corresponded to red points and second two rows are corresponded to yellow points.}
	\label{fig:generalization_examples}
\end{figure}

\subsection{Summary of model and training hyperparameters}

In this section, we provide our model hyperparameters in
Table \ref{table:hyperB} and Table \ref{table:hyperA}.

\begin{table}[h]
\centering
\resizebox{0.70\textwidth}{!}{
\begin{tabular}{@{}rllllll@{}} \toprule
Operation      & spatial dimensions & filter depth \\ \midrule
\textit{style image} & $256 \times 256$ & $3$ \\
Inception-v3 (mixed 6e)   & $17 \times 17$         & $768$            \\
Reduce mean    & $1 \times 1$         & $768$            \\
Bottleneck (matrix multiply)    & $1 \times 1$         & $100$            \\
Matrix multiply    & $1 \times 1$         & $2758$            \\ \midrule
Optimizer              & \multicolumn{6}{@{}l@{}}{Adam ($\alpha = 0.001$, $\beta_1 = 0.9$, $\beta_2 = 0.999$)}  \\
Parameter updates      & \multicolumn{6}{@{}l@{}}{4M}                     \\
Batch size             & \multicolumn{6}{@{}l@{}}{8}                         \\
Weight initialization  & \multicolumn{6}{@{}l@{}}{Isotropic gaussian ($\mu = 0$, $\sigma = 0.01$)}  \\ \bottomrule
\end{tabular}}
\vspace{0.2cm}
\caption{\label{tab:architecture} Style prediction network hyperparameters.}
\label{table:hyperB}
\end{table}

\begin{table}[h]
\centering
\resizebox{0.85\textwidth}{!}{
\begin{tabular}{@{}rllllll@{}} \toprule
Operation      & Kernel size & Stride & Feature maps & Padding & Nonlinearity \\ \midrule
{\bf Network} -- $256 \times 256 \times 3$ input                              \\
Convolution    & $9$         & $1$    & $32$         &  SAME   & ReLU         \\
Convolution    & $3$         & $2$    & $64$         &  SAME   & ReLU         \\
Convolution    & $3$         & $2$    & $128$        &  SAME   & ReLU         \\
Residual block &             &        & $128$        &         &              \\
Residual block &             &        & $128$        &         &              \\
Residual block &             &        & $128$        &         &              \\
Residual block &             &        & $128$        &         &              \\
Residual block &             &        & $128$        &         &              \\
Upsampling     &             &        & $64$         &         &              \\
Upsampling     &             &        & $32$         &         &              \\
Convolution    & $9$         & $1$    & $3$          &  SAME   & Sigmoid      \\
{\bf Residual block} -- $C$ feature maps                                      \\
Convolution    & $3$         & $1$    & $C$          &  SAME   & ReLU         \\
Convolution    & $3$         & $1$    & $C$          &  SAME   & Linear       \\
               & \multicolumn{6}{@{}l@{}}{\em Add the input and the output}   \\
{\bf Upsampling} -- $C$ feature maps                                          \\
               & \multicolumn{6}{@{}l@{}}{\em Nearest-neighbor interpolation, factor 2} \\
Convolution    & $3$         & $1$    & $C$        &  SAME   & ReLU           \\ \midrule
Padding mode           & \multicolumn{6}{@{}l@{}}{REFLECT}                    \\
Normalization          & \multicolumn{6}{@{}l@{}}{Conditional instance normalization after every convolution} \\
Optimizer              & \multicolumn{6}{@{}l@{}}{Adam ($\alpha = 0.001$, $\beta_1 = 0.9$, $\beta_2 = 0.999$)}  \\
Parameter updates      & \multicolumn{6}{@{}l@{}}{4M}                     \\
Batch size             & \multicolumn{6}{@{}l@{}}{8}                         \\
Weight initialization  & \multicolumn{6}{@{}l@{}}{Isotropic gaussian ($\mu = 0$, $\sigma = 0.01$)}  \\ \bottomrule
\end{tabular}}
\vspace{0.2cm}
\caption{\label{tab:architecture} Style transfer network hyperparameters.}
\label{table:hyperA}
\end{table}

\end{document}